\documentclass[review]{elsarticle}
 
\usepackage[margin=1in]{geometry}
\usepackage{lineno, hyperref}
\modulolinenumbers[5]
\usepackage{times}
\usepackage{epsfig}
\usepackage{graphicx}
\usepackage{amsmath}
\usepackage{amssymb}
\usepackage{tablefootnote}
\usepackage{algorithm}%
\usepackage{algpseudocode}%
\usepackage{multirow}
\usepackage{CJK}
\newcommand{\tabincell}[2]{\begin{tabular}{@{}#1@{}}#2
\end{tabular}}
\usepackage{subfig}

\usepackage{color}
\usepackage[tableposition=top]{caption} 
\usepackage{xtab,afterpage,caption}
\usepackage{framed}

\journal{Computer Vision and Image Understanding}

\begin{document}

\begin{frontmatter}

\title{RGB-D-based Human Motion Recognition with Deep Learning: A Survey}

\author[mymainaddress,mysecondaddress]{Pichao Wang}
\ead{pw212@uowmail.edu.au}

\author[mymainaddress]{Wanqing Li}
\ead{wanqing@uow.edu.au}

\author[mymainaddress]{Philip Ogunbona}
\ead{philipo@uow.edu.au}

\author[mythirdaddress]{Jun Wan\corref{mycorrespondingauthor}}
\cortext[mycorrespondingauthor]{Corresponding author}
\ead{jun.wan@nlpr.ia.ac.cn}

\author[myfourthaddress]{Sergio Escalera}
\ead{sergio@maia.ub.es}

\address[mymainaddress]{Advanced Multimedia Research Lab, University of Wollongong, Australia}
\address[mysecondaddress]{Motovis Inc., Adelaide, Australia}
\address[mythirdaddress]{Center for Biometrics and Security Research (CBSR)\& National Laboratory of Pattern Recognition (NLPR), Institute of Automation, Chinese Academy of Sciences (CASIA), Beijing, China}
\address[myfourthaddress]{University of Barcelona and Computer Vision Center, Campus UAB, Barcelona, Spain}

\begin{abstract}
Human motion recognition is one of the most important branches of
human-centered research activities. In recent years, motion recognition 
based on RGB-D data has attracted much attention.  Along 
with the development in artificial intelligence, deep learning techniques have 
gained remarkable success in computer vision. In particular, convolutional 
neural networks (CNN) have achieved great success for image-based tasks, and 
recurrent neural networks (RNN) are renowned for sequence-based problems. 
Specifically, deep learning methods based on the CNN and RNN architectures have 
been adopted for motion recognition using RGB-D data. In this paper, a detailed 
overview of recent advances in RGB-D-based motion recognition is presented. 
The reviewed methods are broadly categorized into four groups, depending on the 
modality adopted for recognition: RGB-based, depth-based, skeleton-based and 
RGB+D-based. As a survey focused on the application of deep 
learning to RGB-D-based motion recognition, we explicitly discuss the 
advantages and limitations of existing techniques. Particularly, we highlighted 
the methods of encoding spatial-temporal-structural information 
inherent in video sequence, and discuss potential directions for future 
research. 
\end{abstract}


\end{frontmatter}


\section{Introduction}
Among the several human-centered research activities (e.g. human 
detection, tracking, pose estimation and motion recognition) in computer vision, 
human motion recognition is particularly important due to its potential 
application in video surveillance, human computer interfaces, ambient 
assisted living, human-robot interaction, intelligent driving, etc. A human 
motion recognition task can be summarised as the automatic 
identification of human behaviours from images or video sequences. The 
complexity and duration of the motion involved can be used as basis for 
broad categorization into four kinds namely gesture, action, interaction and 
group activity. A \textit{gesture} can be defined as the 
basic movement or positioning of the hand, arm, body, or head that 
communicates an idea, emotion, etc. ``Hand waving" and ``nodding"  are 
some typical examples of gestures. Usually, a gesture has relatively short 
duration. An \textit{action} is considered as a type 
of motion performed by a single person during  short time period 
and  involves multiple body parts, in contrast with the few body parts that involved in gesture.  An \textit{activity} is composed by a sequence of actions.
An \textit{interaction} is a type of motion performed 
by two actors; one actor is human while the other may be human or an 
object. This implies that the interaction category will include human-human or 
human-object interaction. ``Hugging each other" and ``playing guitar" are 
examples of these two kinds of interaction, respectively. \textit{Group 
activity} is the most complex type of activity, and it may be a combination of  
gestures, actions and interactions. Necessarily, it involves more than two 
humans and from zero to multiple objects. Examples of group activities 
would include ``two teams playing basketball" and ``group meeting".

Early research on human motion recognition was dominated by the analysis of still images or videos~\citep{aggarwal1999human,wang2003recent,turaga2008machine,
poppe2010survey,guo2014survey,zhu2016handcrafted}.  
Most of these efforts used color and texture cues in 2D images for 
recognition. However, the task remains challenging due to problems posed by 
background clutter, partial occlusion, view-point, lighting changes, execution 
rate and biometric variation. This challenge remains even with current 
deep learning approaches~\citep{herath2017going,FG2017}. 

With the recent development of cost-effective RGB-D sensors, such as Microsoft 
Kinect~\texttrademark and Asus Xtion~\texttrademark, RGB-D-based motion 
recognition has attracted much attention. This is largely because the 
extra dimension (depth) is insensitive to illumination changes and includes rich 
3D structural information of the scene. Additionally, 3D positions of body 
joints can be estimated from depth maps~\citep{Shotton2011}. As a consequence, 
several methods based on RGB-D data have been proposed and the approach 
has proven to be a promising direction for human motion analysis. 

Several survey papers have summarized the research on human motion recognition 
using RGB-D 
data~\citep{chen2013survey,ye2013survey,aggarwal2014human,cheng2016survey,
zhang2016rgb,escalera2016challenges,presti20163d,han2017space}. Specifically, 
Chen et al.~\citep{chen2013survey} focused on depth sensors, 
pre-processing of depth data, depth-based action recognition methods and 
datasets. In their work, Ye et al.~\citep{ye2013survey} presented an overview 
of approaches using depth and skeleton modalities for tasks
including activity recognition, head/hand pose estimation, facial feature 
detection and gesture recognition. The survey presented 
by Aggarwal and Xia~\citep{aggarwal2014human} 
summarized five categories of representations based on 3D silhouettes, skeletal 
joints/body part location, local spatial-temporal features, scene flow features 
and local occupancy features. The work of Cheng et al.~\citep{cheng2016survey} 
focused on RGB-D-based hand gesture recognition datasets and summarized 
corresponding methods from three perspectives: static hand gesture recognition, 
hand trajectory gesture recognition and continuous hand gesture recognition. 
In another effort Escalera et al.~\citep{escalera2016challenges} 
reviewed the challenges and methods for gesture recognition using multimodal 
data.  Some of the surveys have focused on available datasets for 
RGB-D research. For example, the work of Zhang et al.~\citep{zhang2016rgb} 
described available benchmark RGB-D datasets for action/activity 
recognition and included 27 single-view datasets, 10 multi-view 
datasets and 7 multi-person datasets. Other works as Presti and La Cascia~\citep{presti20163d} 
and Han et al.~\citep{han2017space} mainly reviewed skeleton-based 
representation and approaches for action recognition. A short survey on RGB-D action recognition using deep learning was recently presented in~\citep{FG2017}, analysing RGB and depth cues in terms of 2DCNN, 3DCNN, and Deep temporal approaches


\begin{figure*}[t]
\begin{center}
{\includegraphics[height = 55mm, width = 170mm]{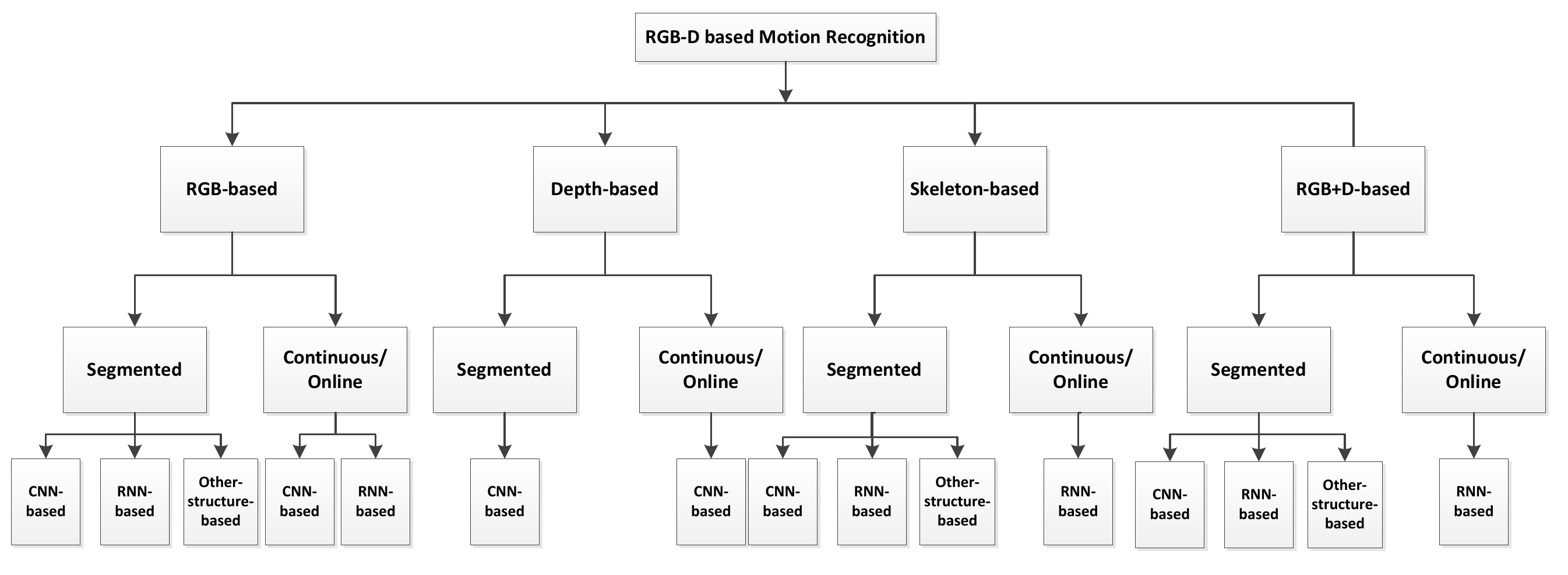}}
\end{center}
\caption{Categorisation of the methods for 
RGB-D-based motion recognition using deep learning.}
\label{categorization}
\end{figure*}

All above surveys mainly focused on the analysis of handcrafted features. Here, we provide a comprehensive review of RGB-D-based human motion recognition using deep learning approaches. Even while focusing on deep learning approaches, the nature of the input 
data is still important. RGB-D data for human motion analysis comprises
three modalities: RGB, depth and skeleton. The main characteristic of RGB data 
is its shape, color and texture which brings the benefits of extracting 
interesting points and optical flow. Compared to RGB 
videos, the depth modality is insensitive to illumination variations, invariant 
to color and texture changes, reliable for estimating body silhouette and 
skeleton, and provides rich 3D structural information of the scene. Differently 
from RGB and depth, skeleton data containing the positions of human joints, is 
a relatively high-level feature for motion recognition. 
The different properties of the three modalities have inspired the 
various methods found in the literature. For example, optical 
flow-based methods with Convolutional Neural Networks (CNN) is very effective 
for RGB channel~\citep{duan2016multi}; depth rank pooling based-method with CNN 
is a good choice for depth modality~\citep{wang2016large}; sequence based method 
with Recurrent Neural Networks (RNN)~\citep{junliu2017} and image-based method 
with CNN~\citep{pichao2016} are effective for skeleton; and scene flow-based 
method using CNN are promising for RGB+D channels~\citep{Pichaocvpr2017}. These 
methods are very effective for specific modalities, but not always the case for 
all the modalities. Given these observations, this survey identified four 
broad categories of methods based on the modality adopted for human motion
recognition.  The categories include RGB-based, depth-based, skeleton-based and 
RGB+D-based. 

In each category, two sub-divisions are further identified, namely segmented 
human motion recognition and continuous/online motion recognition.  For 
segmented motion recognition, the scenario of the problem can be simply 
described as classifying a well delineated sequence of video frames as one of a 
set of motion types. This is in contrast to continuous/online human motion 
recognition where there are no a priori given boundaries of motion execution. 
The online situation is compounded by the fact that the video sequence is not 
recorded and the algorithm must deal with frames as they are being captured, 
save for possibly a small data cache.

During the performance of a specified motion spatial information which refers 
to the spatial configuration of human body at an instant of time (e.g. relative 
positions of the human body parts) can be identified. Similarly, there is the 
temporal information which characterizes the spatial configuration of the body 
over time (i.e. the dynamics of the body). Lastly, the structural information 
encodes the coordination and synchronization of body parts over the period in 
which the action is being performed. It describes the relationship of 
the spatial configurations of human body across different time slots.

In reviewing the various methods, consideration has been given to the manner in 
which the spatial, temporal and structural information have been exploited.  
Hence, the survey discusses the advantages and limitations of the reviewed 
methods from the spatial-temporal-structural encoding viewpoint, and suggests 
potential directions for future research.

A key novelty of this survey is the focus on three architectures of neural 
networks used in the various deep learning methods reviewed namely CNN-based, 
RNN-based and other structured networks. Fig.~\ref{categorization} illustrates 
the taxonomy underpinning this survey.  

This is one of the first surveys dedicated 
to RGB-D-based human motion recognition using deep learning. Apart from 
this claim, this survey distinguishes itself from other surveys through 
the following contributions: 
\begin{itemize}
\item Comprehensive coverage of the most recent and advanced deep 
learning-based methods 
developed in the last five years, thereby providing readers with a complete 
overview of recent research results and state-of-the-art methods.
\item Insightful categorization and analysis of methods based on the 
different properties of the modalities; highlight of the pros and cons of the 
methods described in the reviewed papers from the viewpoint of 
spatial-temporal-structural encoding.
\item Discussion of the challenges of RGB-D-based motion recognition; analysis 
of the limitations of available methods and discussion of potential research 
directions. 
\end{itemize}

Additionally, several recently released or commonly used RGB-D-based benchmark 
datasets associated with deep learning are surveyed. The main application domain 
of interest in this survey paper is human motion recognition based on RGB-D 
data, including  gesture recognition, 
action/activity recognition and  interaction recognition. The lack of 
datasets focused on RGB-D-based group activity recognition has led to 
paucity of research on this topic and thus this survey does not 
cover this topic. Other RGB-D-based human-centered applications, 
such as human detection, tracking and pose estimation, are also not the focus 
of this paper. For surveys on RGB-D data acquisition readers are 
referred to~\citep{chen2013survey,cheng2016survey,han2017space}.

Subsequent sections of the this survey are organized as 
follows. Commonly used RGB-D-based benchmark datasets are described in 
Section~\ref{dataset}. 
Sections~\ref{rgb} to~\ref{multi} discuss methods of RGB-D-based 
motion recognition using deep learning from four perspectives: RGB-based motion 
recognition, depth-based motion recognition, skeleton-based motion 
recognition and RGB+D-based motion recognition. Challenges 
of RGB-D-based motion 
recognition and pointers to future directions are presented in 
Section~\ref{discuss}. The survey provides concluding remarks in 
Section~\ref{conclusion}.

\section{Benchmark Datasets}\label{dataset}
Over the last decade, a number of RGB-D benchmark 
datasets have been collected and made publicly available for the research 
community. The sources of the datasets are mainly of three 
categories~\citep{chen2013survey,cheng2016survey,han2017space}: Motion 
capture (Mocap) system, structured-light cameras (e.g. Kinect v1) and 
time-of-flight (ToF) cameras (e.g. Kinect v2). Hence the modalities of 
the datasets cover RGB, depth, skeleton and their combinations.  With the advance of deep learning, deep methods have been developed for estimating skeletons directly from single images or video sequences, such as DeepPose~\citep{toshev2014deeppose}, Deepercut~\citep{insafutdinov2016deepercut} and Adversarial PoseNet~\citep{Chen_2017_ICCV}. A 
comprehensive survey of these datasets have appeared in the literature (see 
e.g.~\citep{cheng2016survey} for hand gestures and~\citep{zhang2016rgb} for 
action recognition).  In the present survey only 15 large-scale datasets that 
have been commonly adopted for evaluating deep learning-based methods are 
described. The reader is referred to Table~\ref{performance} for a sample of works 
(publications) that have used these datasets. For the purpose of this survey, 
the datasets have been divided into two groups: segmented datasets 
and continuous/online datasets.

\subsection{Segmented Datasets}

By segmented datasets we refer to those datasets where samples correspond 
to a whole begin-end action/gestures, with one segment for one action. They are 
mainly used for classification purposes. The following are several segmented 
datasets commonly used for the evaluation of methods based on deep 
learning.


\subsubsection{CMU Mocap}
CMU Graphics Lab Motion Capture Database (CMU 
Mocap)~\citep{CMUMocap}(\url{http://mocap.cs.cmu.edu/}) 
is one of the earliest source of data that covers a wide range of human actions, 
including interactions between two subjects, human locomotion, interactions with 
uneven terrain, sports, and other human actions. This dataset consists of RGB and
skeleton modalities.
\subsubsection{HDM05}
Motion Capture Database HDM05~\citep{cg-2007-2} 
(\url{http://resources.mpi-inf.mpg.de/HDM05/}) 
was captured by an optical marker-based technology with the frequency of 
120~Hz, which contains 2337 sequences for 130 actions performed by 
5 non-professional actors, and 31 joints in each frame. Besides skeleton data, 
this dataset also provides RGB data.

\subsubsection{MSR-Action3D}
MSR-Action3D~\citep{li2010action} 
(\url{http://www.uow.edu.au/~wanqing/#MSRAction3DDatasets}) is the first public 
benchmark RGB-D action dataset collected using Kinect~\texttrademark sensor by 
Microsoft Research, Redmond and University of Wollongong in 2010. 
The dataset contains 20 actions: \textit{high arm wave}, \textit{horizontal arm 
wave}, \textit{hammer}, \textit{hand catch}, \textit{forward punch}, 
\textit{high throw}, \textit{draw x}, \textit{draw tick}, \textit{draw circle}, 
\textit{hand clap}, \textit{two hand wave}, \textit{side-boxing}, \textit{bend}, 
\textit{forward kick}, \textit{side kick}, \textit{jogging}, \textit{tennis 
serve}, \textit{golf swing}, \textit{pickup and throw}.  
Ten subjects performed these actions three times. All the 
videos were recorded from a fixed point of view and the subjects were facing the 
camera while performing the actions. The background of the dataset 
was removed by some post-processing. Specifically, if an action needs to be 
performed with one arm or one leg, the actors were required to perform it using 
right arm or leg. 

\subsubsection{MSRC-12}
MSRC-12 
dataset~\citep{Fothergill2011}(\url{
http://research.microsoft.com/en-us/um/cambridge/projects/msrc12/}) 
was collected by Microsoft Research Cambridge and University of Cambridge in 
2012. 
The authors provided three familiar and easy to prepare instruction modalities and 
their combinations to the participants. The modalities are (1) descriptive text 
breaking down the performance kinematics, (2) an ordered series of static 
images of a person performing the gesture with arrows annotating as appropriate, 
and (3) video (dynamic images) of a person performing the gesture. 
There are 30 participants in total and for each 
gesture, the data were collected as: Text 
(10 people), Images (10 people), Video (10 people), Video with text (10 people), 
Images with text (10 people). The dataset was captured using 
one Kinect~\texttrademark sensor and only the skeleton data are made available.

\subsubsection{MSRDailyActivity3D}
MSRDailyActivity3D 
Dataset~\citep{wang2012mining}(\url{
http://www.uow.edu.au/~wanqing/#MSRAction3DDatasets}) 
was collected by  Microsoft and the Northwestern University in 2012 and focused 
on daily activities. The motivation was to cover human daily activities in the 
living room. 
The actions were performed by 10 actors while sitting on 
the sofa or standing close to the sofa. The camera was fixed in front of the 
sofa. In addition to depth data, skeleton data are also recorded, but the joint 
positions extracted by the tracker are very noisy due to the actors being 
either sitting on or standing close to the sofa. 
\subsubsection{UTKinect}
UTKinect 
dataset~\citep{xia2012view}(\url{
http://cvrc.ece.utexas.edu/KinectDatasets/HOJ3D.html}) 
was collected by the University of Texas at 
Austin in 2012. Ten types of human actions were performed twice by 10 subjects. 
The subjects performed the actions from a variety of 
views. One challenge of the dataset is due to the actions being 
performed with high actor-dependent variability. Furthermore, human-object 
occlusions and body parts being out of the field of view have further increased the difficulty 
of the dataset. 
Ground truth in terms of action labels and 
segmentation of sequences are provided. 
\subsubsection{G3D}
Gaming 3D dataset (G3D)~\citep{bloom2012g3d}(\url{http://dipersec.king.ac.uk/G3D/}) captured by 
Kingston University in 2012 focuses on real-time action recognition in gaming 
scenario. 
It contains 10 subjects performing 20 gaming actions.
Each subject performed these actions thrice.
Two kinds of labels were provided as ground truth: the onset and 
offset of each action and the peak frame of each action.

\subsubsection{SBU Kinect Interaction Dataset}
SBU Kinect Interaction Dataset~\citep{yun2012two}(\url{http://www3.cs.stonybrook.edu/~kyun/research/kinect_interaction/index.html}
) was collected by Stony Brook University in 2012. It contains 
eight types of interactions. 
All videos were recorded with the same indoor background. Seven participants were 
involved in performing the activities which have interactions 
between two actors. The dataset is segmented into 21 sets and each set contains 
one or two sequences of each action category. Two kinds of ground truth 
information are provided: action labels of each segmented video and 
identification of ``active'' actor and ``inactive'' actor.


\subsubsection{Berkeley MHAD}
Berkeley Multimodal Human Action Database (Berkeley 
MHAD)~\citep{ofli2013berkeley}(\url{http://tele-immersion.citris-uc.org/berkeley_mhad#dl}), 
collected by University of California at Berkeley and Johns Hopkins University 
in 2013, was captured in five different modalities to expand the fields of 
application. The modalities are derived from: optical mocap system, four 
multi-view stereo vision cameras, two Microsoft Kinect v1 cameras, six wireless 
accelerometers and four microphones. Twelve subjects performed 11 actions, 
five times each. Three categories of actions are included: (1) actions with 
movement in full body parts, e.g., \textit{jumping in place}, \textit{jumping 
jacks}, \textit{throwing}, etc., (2) actions with high dynamics in upper 
extremities, e.g.,\textit{ waving hands}, \textit{clapping hands}, etc. and (3) 
actions with high dynamics in lower extremities, e.g., \textit{sit down}, 
\textit{stand up}. The actions were executed with style and speed variations. 

\subsubsection{Northwestern-UCLA Multiview Action 3D}
Northwestern-UCLA Multiview Action 
3D~\citep{wang2014cross}(\url{
http://users.eecs.northwestern.edu/~jwa368/my_data.html})
was collected by Northwestern University and  University of California at Los 
Angles in 2014. This dataset contains data taken from a variety of viewpoints. 
The actions were performed by 10 actors and captured by 
three simultaneous Kinect\texttrademark v1 cameras. 


\subsubsection{ChaLearn LAP IsoGD}ChaLearn LAP IsoGD 
Dataset~\citep{wanchalearn}~(\url{http://www.cbsr.ia.ac.cn/users/jwan/database/isogd.html}) 
is a large RGB-D dataset for segmented gesture recognition, and it was collected 
by Kinect v1 camera. It includes 47933 RGB-D depth sequences, each RGB-D video 
representing one gesture instance. 
There are 249 gestures performed by 21 different individuals. The dataset is 
divided into training, validation and test sets. All three sets consist of 
samples of different subjects to ensure that the gestures of one subject in the
validation and test sets will not appear in the training set.

\subsubsection{NTU RGB+D}
NTU RGB+D 
Dataset~\citep{shahroudy2016ntu}(\url{https://github.com/shahroudy/NTURGB-D }) 
is currently the largest action recognition dataset in terms of the number of 
samples per action. The RGB-D data is captured by Kinect v2 cameras. The dataset 
has more than 56 thousand sequences and 4 million frames, containing 60 actions 
performed by 40 subjects aging between 10 and 35. It consists of front view, two 
side views and left, right 45 degree views. 

\subsection{Continuous/Online Datasets}
Continuous/online datasets refer to those datasets 
where each video sequence may contain one or more actions/gestures, and the 
segmented position between different motion classes are unknown. These datasets 
are mainly used for action detection, localization and online prediction. There 
are few datasets for this type.

\subsubsection{ChaLearn2014 Multimodal Gesture Recognition}
ChaLearn2014 Multimodal Gesture 
Recognition~\citep{escalera2014chalearn}~(\url{http://gesture.chalearn.org/2014-looking-at-people-challenge}) 
is multi-modal dataset collected by Kinect v1 sensor, including RGB, depth, 
skeleton and audio modalities. In all sequences, a single user is recorded in 
front of the camera, performing natural communicative Italian gestures.  The 
starting and ending frames for each gesture are annotated along with
the gesture class label.
It contains nearly 14K manually labeled (beginning and ending frame) gesture 
performances in continuous video sequences, with a vocabulary of 20 Italian 
gesture categories.
There are 1, 720, 800 labeled frames across 13, 858
video fragments of about 1 to 2 minutes sampled at 20~Hz. The gestures are 
performed by 27 different individuals under diverse conditions; these
include varying clothes, positions, backgrounds and lighting.

\begin{footnotesize}
\begin{table*}
\caption{Statistics of the public available benchmark datasets that are commonly 
used for evaluation with deep learning. 
Notation for the header: Seg: Segmented, Con:Continuous, D: Depth, S:Skeleton, Au:Audio, Ac:Accelerometer, IR:IR videos, 
\#:number of, JI:Jaccard Index. \label{dataset1}}
\begin{tabular}{|c|c|c|c|c|c|c|c|c|c|}
\hline
Dataset & year & \tabincell{c}{Acquisition\\device}& Seg/Con   & Modality   & \#Class  & \#Subjects  & \#Samples &\#Views & Metric\\\hline                                
CMU Mocap  & 2001 & Mocap & Seg   & RGB,S   & 45  & 144 &2,235 & 1 & Accuracy\\ \hline                                     
HDM05 & 2007 & Mocap &Seg & RGB,S& 130 &  5  & 2337 & 1 & Accuracy\\\hline                                 
MSR-Action3D & 2010 & Kinect v1 &Seg  & S,D & 20   &  10   & 567 & 1& Accuracy\\\hline                                     
MSRC-12  & 2012 & Kinect v1 & Seg  & S & 12 & 30  & 594 & 1& Accuracy\\\hline
\tabincell{c}{MSR\\DailyActivity3D}  & 2012 & Kinect v1 &Seg   &RGB,D,S   & 16 & 10 & 320 & 1 & Accuracy\\\hline
UTKinect  & 2012 & Kinect v1 &Seg  &  RGB,D,S   & 10   &  10 & 200 &1 &Accuracy\\\hline
G3D  & 2012 & Kinect v1 &Seg  & RGB,D,S   & 5 & 5 & 200 & 1 & Accuracy\\\hline
\tabincell{c}{SBU Kinect\\Interaction}  &  2012  & Kinect v1  & Seg & RGB,D,S &7 & 8   & 300  &1 &Accuracy\\\hline
Berkeley MHAD  & 2013 & \tabincell{c}{Mocap\\Kinect v1} & Seg &RGB,D,S,Au,Ac   & 12 &  12  & 660 & 4 & Accuracy\\\hline
\tabincell{c}{Multiview\\Action3D} & 2014 & Kinect v1 & Seg   & RGB,D,S & 10 & 10 & 1475 & 3 & Accuracy \\\hline
\tabincell{c}{ChaLearn LAP\\IsoGD} & 2016 & Kinect v1 &Seg  & RGB,D  & 249  & 21 & 47933 &1 & Accuracy\\\hline
NTU RGB+D & 2016 &Kinect v2 & Seg  & RGB,D,S,IR &  60 &  40  & 56880 & 80 & Accuracy\\\hline
ChaLearn2014 & 2014  & Kinect v1 & Con   & RGB,D,S,Au & 20 & 27  & 13858 & 1 & \tabincell{c}{Accuracy\\JI etc.}\\\hline
\tabincell{c}{ChaLearn LAP\\ ConGD} & 2016 & Kinect v1 & Con &  RGB,D& 249 & 21 & 22535 & 1 &JI\\\hline  
PKU-MMD & 2017 &  Kinect v2 &Con & RGB,D,S,IR & 51 & 66 & 1076 
& 3 & JI etc. \\                                                        
\hline  
\end{tabular}
\end{table*}
\end{footnotesize}

\subsubsection{ChaLearn LAP ConGD} The ChaLearn LAP ConGD 
Dataset~\citep{wanchalearn}~(\url{http://www.cbsr.ia.ac.cn/users/jwan/database/congd.html}) 
is a large RGB-D  dataset for continuous gesture recognition. It was collected 
by Kinect v1 sensor and includes 47933 RGB-D  gesture instances in 22535 RGB-D 
gesture videos. Each RGB-D 
video may contain one or more gestures.  There are 249 gestures performed by 21 
different individuals. The dataset is 
divided into training, validation and test sets. All three sets consist of 
samples of different subjects to ensure that the gestures of one subject in the
validation and test sets will not appear in the training set.

\subsubsection{PKU-MMD}
PKU-MMD~\citep{liu2017pku}~(\url{http://www.icst.pku.edu.cn/struct/Projects/PKUMMD.html}) 
is a large scale dataset for continuous multi-modality 3D human action 
understanding and covers a wide range of complex human activities with well 
annotated information. It was captured via the Kinect v2 sensor. PKU-MMD 
contains 1076 long video sequences in 51 action categories, performed by 66 
subjects in three camera views. It contains almost 20,000 action instances and 
5.4 million frames in total. It provides multi-modality data sources, including 
RGB, depth, Infrared Radiation and Skeleton.  

\begin{figure*}[hp]
\begin{center}
{\includegraphics[height = 240mm, width = 165mm]{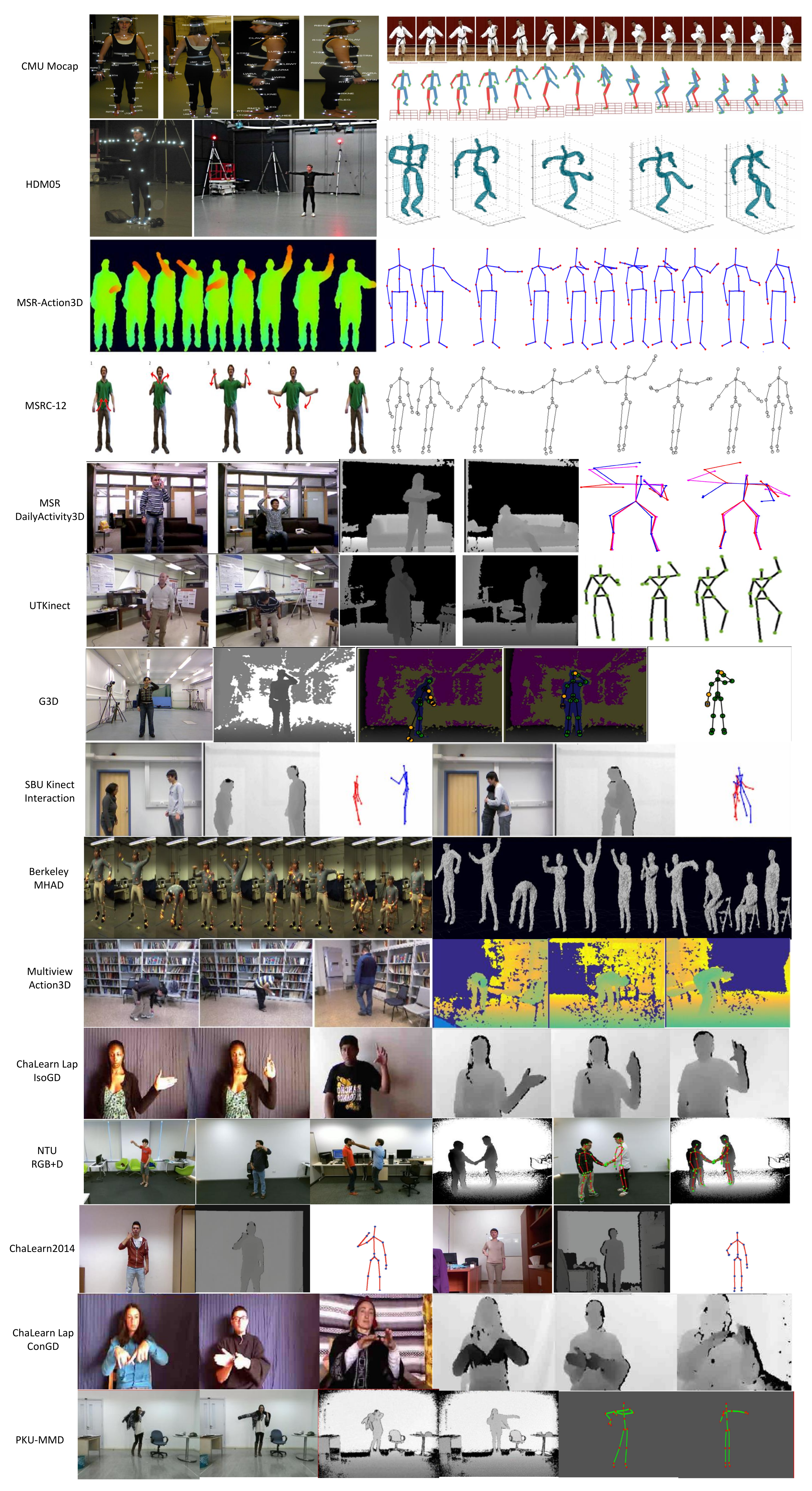}}
\end{center}
\caption{Sample images from different datasets.}
\label{datasetsample}
\end{figure*}

Table~\ref{dataset} shows the statistics of publicly available benchmark 
datasets that are commonly used for evaluation of deep learning-based 
algorithms. It can be seen that the surveyed datasets cover a wide range of 
different types of actions including gestures, simple actions, daily 
activities, human-object interactions, human-human interactions. It also covers 
both segmented and continuous/online datasets, with different acquisition 
devices, modalities, and views. Sample images from different datasets are shown in Fig.~\ref{datasetsample}.

In this section, we introduce the deep learning concepts and architectures 
that are relevant or have been applied to RGB-D-based motion recognition. 
Readers who are interested in more background and techniques are referred to  
the book by~\citep{Goodfellow-et-al-2016}.

\section{RGB-based Motion Recognition with Deep Learning}\label{rgb}

RGB is one important channel of RGB-D data with characteristics 
including shape, color and texture that bear rich features.  These properties 
also make it effective to directly use networks, such as 2D 
CNNs~\citep{krizhevsky2012imagenet,simonyan2014very,he2016deep}, to extract 
frame-level features. Even though most of the surveyed methods for 
this section are not adapted to RGB-D-based datasets, we argue that the 
following methods could be directly adapted to RGB modality of RGB-D datasets. 
We define three categories namely, CNN-based, RNN-based and 
other-architecture-based approaches for segmented motion 
recognition; the first two categories are for continuous/online motion 
recognition.

\subsection{Segmented Motion Recognition}

\subsubsection{CNN-based Approach} 
\begin{figure}[t]
\begin{center}
{\includegraphics[scale=1]{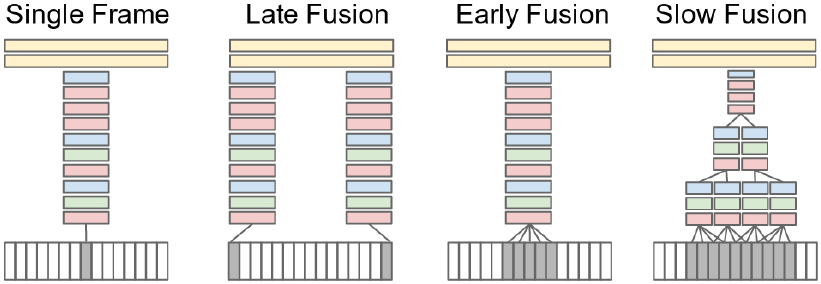}}
\end{center}
\caption{Different approaches for fusing information over temporal dimension 
through the network.  Red, green and blue boxes indicate convolutional, 
normalization and pooling layers respectively.  In the Slow Fusion model, the 
depicted columns share parameters. Figure from~\citep{karpathy2014large}.}
\label{CNNfusion}
\end{figure}

For this group of methods, currently 
there are mainly four approaches to encode spatial-temporal-structural 
information. The first approach applies CNN to extract features from individual 
frames and later, fuse the temporal information. For 
example~\citep{karpathy2014large} investigated four temporal fusion methods, and 
proposed the concept of slow fusion where higher layers get access to 
progressively more global information in both spatial and temporal dimensions (see Fig.~\ref{CNNfusion}).
Furthermore, several temporal pooling methods have been explored and 
the suggestion is that max pooling in the temporal domain is 
preferable~\citep{yue2015beyond}.

\begin{figure*}
\begin{center}
{\includegraphics[height = 12mm, width = 165mm]{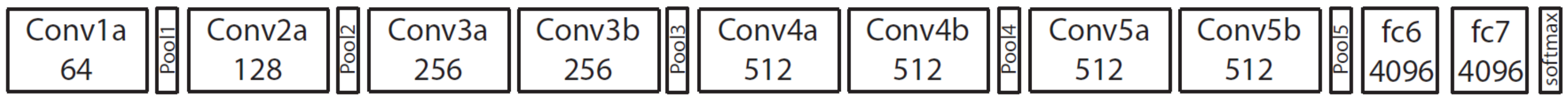}}
\end{center}
\caption{C3D net has 8 convolutions, 5 max-poolings, and 2 fully connected 
layers, followed by a softmax output layer. All 3D convolution kernels are 
3$\times$3$\times$3 with stride 1 in both spatial and temporal dimensions. 
Number of filters are indicated in each box. The 3D pooling layers are 
as indicated from \textit{pool1} to \textit{pool5}. All pooling kernels are 
2$\times$2$\times$2, except for \textit{pool1} which is 1$\times$2$\times$2. 
Each fully connected layer has 4096 output units. Figure 
from~\citep{tran2015learning}.}
\label{c3d}
\end{figure*}

The second approach extends convolutional operation into temporal domain. 
In one such implementation, Ji et al.~\citep{ji20133d} proposed  
3D-convolutional networks 
using 3D kernels (filters extended along the time axis) to extract features from 
both spatial and temporal dimensions. This work empirically showed that the 
3D-convolutional networks outperform their 2D frame-based counterparts. With 
modern deep architectures, such as VGG~\citep{simonyan2014very}, and large-scale 
supervised training datasets, such as Sports-1M~\citep{karpathy2014large}, 
Tran et al~\citep{tran2015learning} extended the work presented 
in~\citep{ji20133d} by including 3D pooling layers, and proposed a generic 
descriptor named~\textit{C3D} by 
averaging the outputs of the first fully connected layer of the networks (see 
Fig.~\ref{c3d}). However, both of these works break the video sequence 
into short clips and aggregate video-level information by late score fusion.  
This is likely to be suboptimal when considering some long action 
sequence, such as walking or swimming that lasts several seconds and spans tens 
or hundreds of video frames. To handle this problem, Varol et 
al.~\citep{varol2016long} investigated the learning of long-term video 
representations and proposed the Long-term Temporal Convolutions (LTC) at the 
expense of decreasing spatial resolution to keep the complexity of networks 
tractable. Despite the fact that this is straightforward and mainstream, 
extending spatial kernels to 3D spatio-temporal derivative inevitably 
increases the number of parameters of the network. To relieve the  
drawbacks of 3D kernels, Sun et al.~\citep{sun2015human} factorized a 3D filter 
into a combination of 2D and 1D filters.

\begin{figure}[t]
\begin{center}
{\includegraphics[scale=1.5]{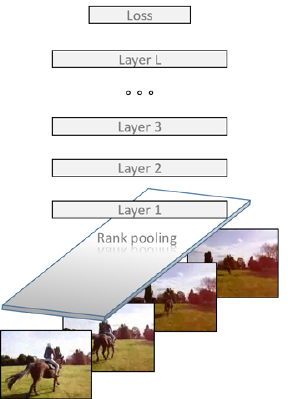}}
\end{center}
\caption{Rank pooling encodes the RGB video into one dynamic image and CNN is
adopted for feature extraction and classification. Figure 
from~\citep{bilen2016dynamic}.}
\label{DI}
\end{figure}

The third approach is to encode the video into dynamic images that contain 
the spatio-temporal information and then apply CNN for image-based recognition. 
Bilen et al.~\citep{bilen2016dynamic} proposed to adopt rank 
pooling~\citep{fernando2016rank} to encode the video into one dynamic set 
of images and used pre-trained models over 
ImageNet~\citep{krizhevsky2012imagenet} for fine-tuning (see Figure.~\ref{DI}). 
The end-to-end learning methods with rank pooling has also been proposed 
in~\citep{bilen2016dynamic,FernandoICML2016}. Hierarchical rank 
pooling~\citep{Fernando2016a} is proposed to learn higher order and non-linear 
representations compared to the original work. Generalized rank 
pooling~\citep{cherian2017generalized} is introduced to improve the original 
method via a quadratic ranking function which jointly provides a low-rank 
approximation to the input data and preserves their temporal order in a 
subspace. 

\begin{figure*}[t]
\begin{center}
{\includegraphics[height = 40mm, width = 165mm]{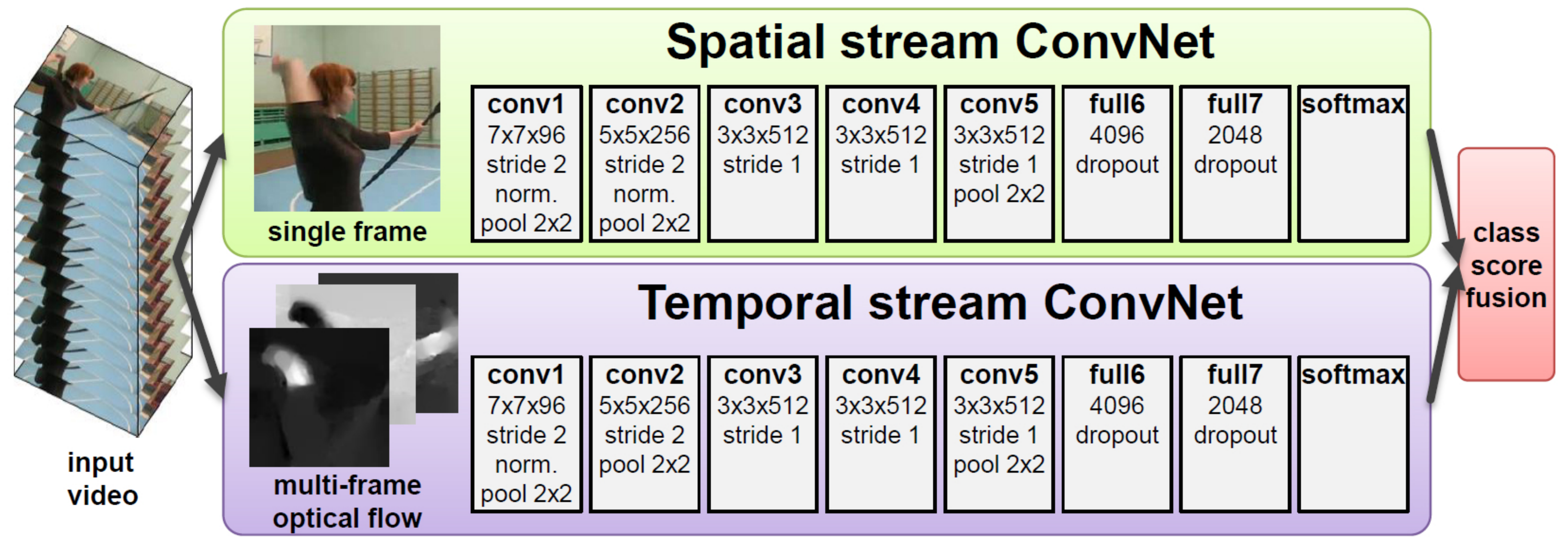}}
\end{center}
\caption{Two-stream architecture for RGB-based motion recognition. Figure 
from~\citep{simonyan2014two}.}
\label{twostream}
\end{figure*}

Besides the above works that aim to adopt one network to exploit both 
 spatio-temporal information contained in the video, the fourth approach 
separates the two factors and adopt multiple stream networks. 
Simonyan et al.~\citep{simonyan2014two}  proposed one spatial stream 
network fed with raw video frames, and one temporal stream network accepting 
optical flow fields as input. The two streams are fused together using the 
softmax scores (see Figure.~\ref{twostream} for the two-stream architecture).  
Wang et al.~\citep{wang2015action} extended the two-stream networks by 
integrating improved trajectories~\citep{wang2013dense}, where 
trajectory-constrained sampling and pooling are used to encode deep features 
learned from deep CNN architecture, into effective descriptors. To incorporate 
long-range temporal structure using the two-stream networks,  
Wang et al.~\citep{wang2016temporal} devised a temporal segment network (TSN) 
that uses a sparse sampling scheme to extract short snippets over a long video 
sequence. 
With the removal of redundancy from consecutive frames and a segmental 
structure, aggregated information is obtained from the sampled snippets. To 
reduce the expensive calculation of optical flow, Zhang et 
al.~\citep{zhang2016real} accelerated this two stream structure by replacing 
optical flow with motion vector which can be obtained directly from compressed 
videos without extra calculation. 
Wang et al.~\citep{wang2016two} leveraged semantic cues in video by using  
a two-stream semantic region-based CNNs (SR-CNNs) to incorporate human/object 
detection results into the framework. In their work, 
Ch{\'e}ron et al.~\citep{cheron2015p}
exploit spatial structure of the human pose and extract a pose-based 
convolutional neural network (P-CNN) feature from both RGB frames and optical 
flow for fine-grained action recognition. The work 
presented in~\citep{wang2016actions} 
formulated the problem of action recognition from a new perspective and model 
an action as a transformation which changes the state of the environment before 
the action to the state after the action. They designed a Siamese 
network which models the action as a transformation on a high-level feature 
space based on the two-stream model. Based on the two-stream framework, 
\citep{zhu2016key} proposed a key volume mining deep framework for action 
recognition, where they identified key volumes and conducted classification 
simultaneously. Inspired by the success of Residual Networks 
(ResNets)~\citep{he2016deep}, 
Feichtenhofer et al.~\citep{feichtenhofer2016spatiotemporal} injected 
residual connections between the two streams to allow spatial-temporal 
interaction between them. Instead of using optical flow for 
temporal stream, \citep{lea2016segmental} adopted Motion History 
Image (MHI)~\citep{bobick2001recognition} as the motion clue. The MHI 
was combined with RGB frames in a spatio-temporal CNN for fined 
grained action recognition.  However, all the methods reviewed 
above incorporated the two streams from separate training regimes; any 
 registration of the two streams was neglected. In order to address this 
gap and propose a new architecture for 
spatial-temporal fusion of the two streams 
Feichtenhofer et al.~\citep{feichtenhofer2016convolutional} investigated three 
aspects of fusion for the two streams: (i) how to fuse the two networks with 
consideration for spatial registration, (ii) where to fuse the two networks and, 
(iii) how to fuse the networks temporally. One of their conclusions was that the
results suggest the importance of learning correspondences between
highly abstract ConvNet features both spatially and temporally.

\subsubsection{RNN-based Approach} 
\begin{figure}[t]
\begin{center}
{\includegraphics[scale=0.2]{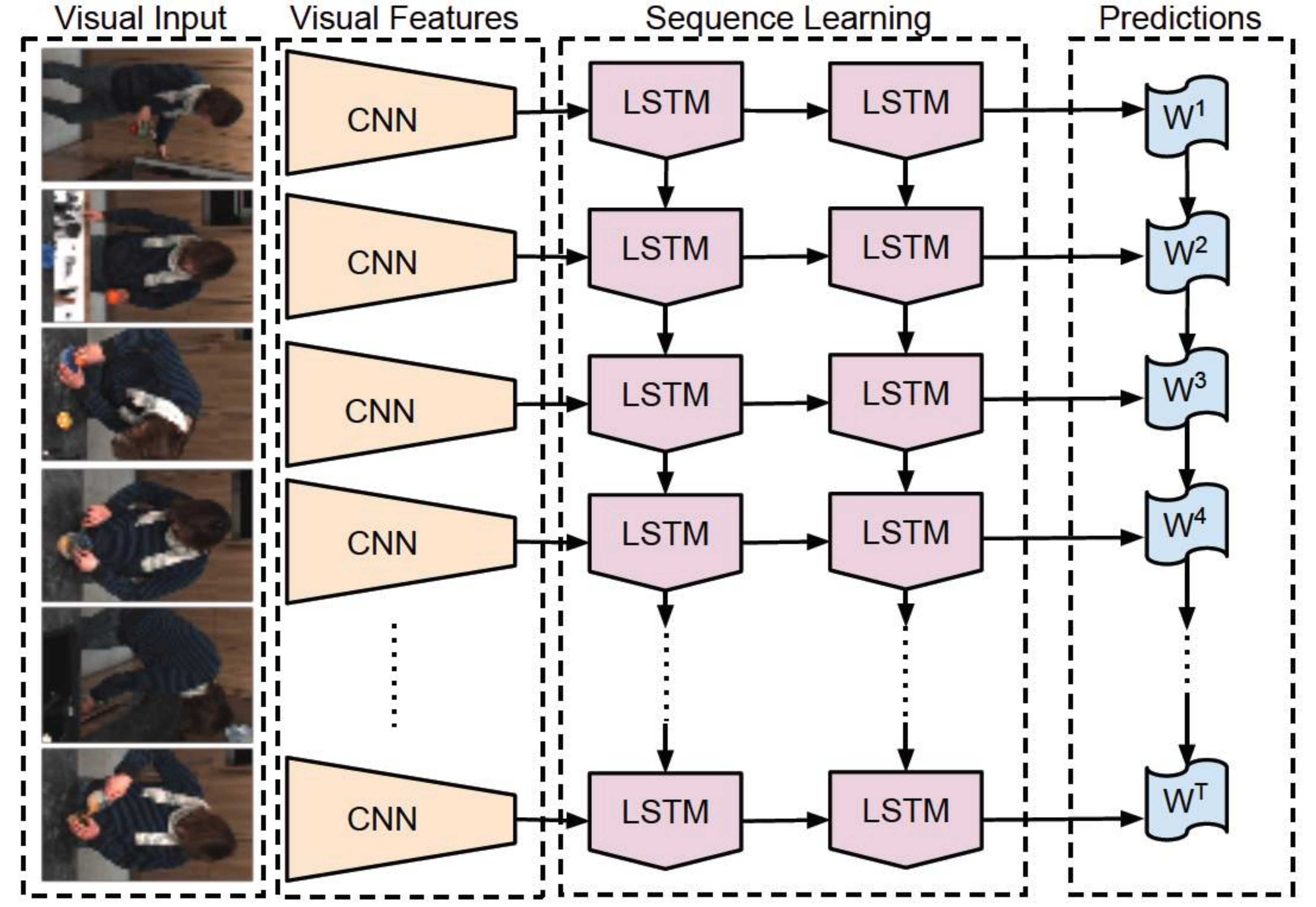}}
\end{center}
\caption{LRCN processes the variable-length visual input with a CNN, 
whose outputs are fed into a stack of recurrent sequence models. The 
output is a variable-length prediction. Figure 
from~\citep{donahue2015long}.}
\label{LRCN}
\end{figure}

\begin{figure*}[t]
\begin{center}
{\includegraphics[height = 60mm, width = 165mm]{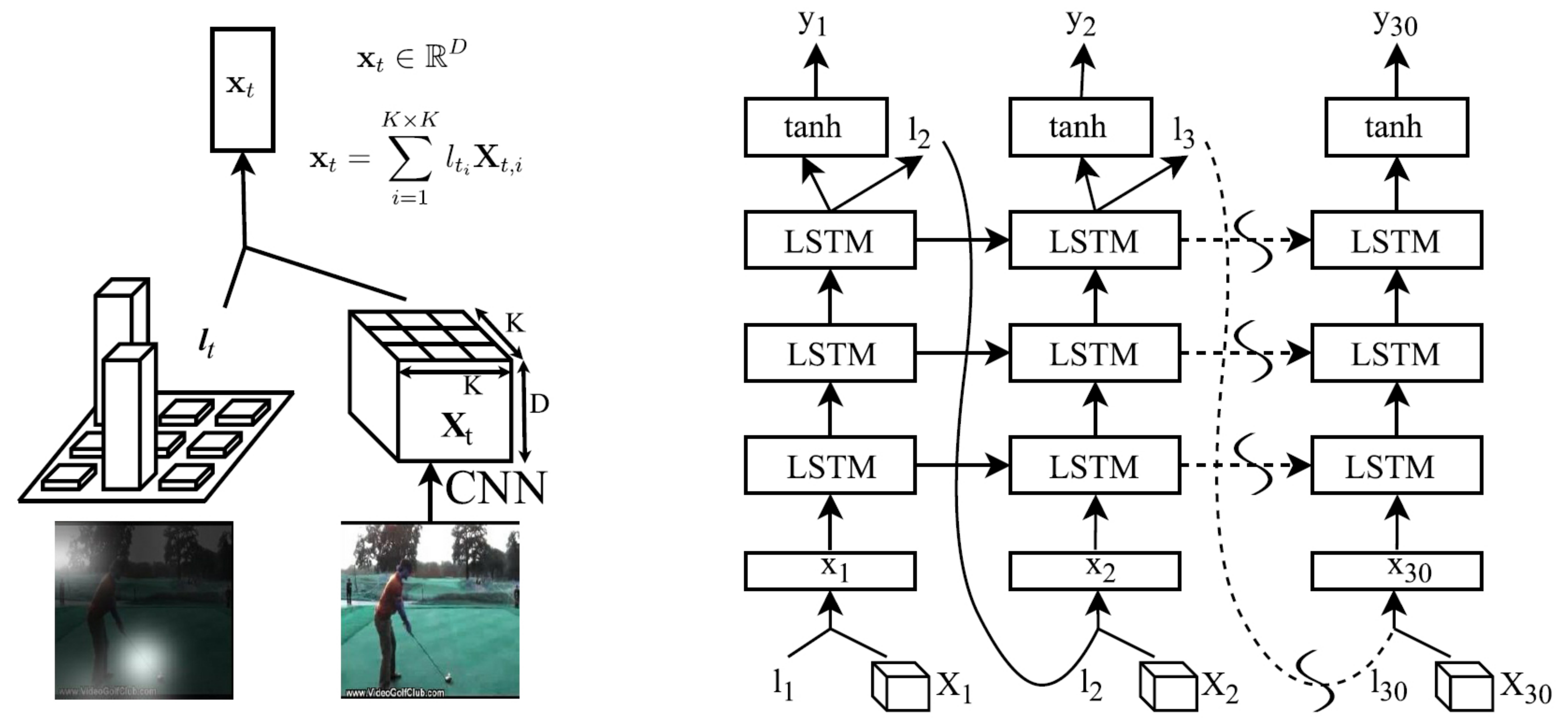}}
\end{center}
\caption{The CNN takes the video frames as its input and produces a feature 
tube. The model computes the current input $\boldsymbol{x}_{t}$ as an average of 
the feature slices weighted according to the location softmax 
$\boldsymbol{I}_{t}$. At each time-step $t$, the recurrent network takes a 
feature slice $\boldsymbol{x}_{t}$ as input. It then propagates 
$\boldsymbol{x}_{t}$ through three layers of LSTMs and predicts the next 
location probabilities $\boldsymbol{I}_{t+1}$ and the class label 
$\boldsymbol{y}_{t}$. Figure from~\citep{sharma2015action}.}
\label{attention}
\end{figure*}

For RNN-based approach, 
Baccouche et al.~\citep{baccouche2011sequential} tackled the problem of action 
recognition through a cascade of 3D CNN and LSTM, in which the two networks 
were trained separately. Differently from the separate training, 
Donahue et al.~\citep{donahue2015long} proposed one Long-term Recurrent 
Convolutional Network (LRCN) to exploit end-to-end training of the two 
networks( see illustration in Figure.~\ref{LRCN}).  To take full advantage of 
both CNN and RNN, Ng et al.~\citep{yue2015beyond} aggregated CNN features  with 
both temporal 
pooling and LSTM for temporal exploitation, and fused the output scores from 
the feature pooling and LSTM network to conduct final action recognition.  
Pigou et al.~\citep{pigou2015beyond} proposed an end-to-end trainable neural 
network architecture incorporating temporal convolutions and bidirectional LSTM 
for gesture recognition. This provided opportunity to mine temporal 
information that is much discriminative for gesture 
recognition. In their work, Sharma et al.~\citep{sharma2015action} proposed a 
soft attention model for action recognition based on LSTM (see 
Figure.~\ref{attention}). The attention model 
learns the parts in the frames that are relevant for the task at hand and 
attaches higher importance to them. Previous 
attention-based methods have only utilized video-level category as supervision 
to train LSTM. This strategy may lack a detailed and dynamical guidance
and consequently restrict their capacity for modelling complex motions in 
videos.  Du et al.~\citep{du2017rpan} address this problem by proposing a 
recurrent pose-attention network (RPAN) for action recognition
in videos, which can adaptively learn a highly discriminative
pose-related feature for every-step action
prediction of LSTM. To take advantage of both Fisher 
Vector~\citep{sanchez2013image} and RNN, Lev et al.~\citep{lev2016rnn} 
introduced a Recurrent Neural Network Fisher Vector (RNN-FV) where the GMM 
probabilistic model in the fisher vector is replaced by a RNN and thus 
avoids the need for the assumptions of data distribution in the GMM. 
Even though RNN is remarkably capable of modeling temporal dependences, 
it lacks an intuitive high-level spatial-temporal structure. The 
spatio-temporal-structural information has been mined by Jain et 
al.~\citep{jain2016structural} through a combination of the powers of 
spatio-temporal graphs and RNN for action recognition.   Recently, 
Sun et al.~\citep{sun2017lattice} proposed a Lattice-LSTM (L2STM) network, which
extends LSTM by learning independent hidden state transitions of memory cells 
for individual spatial locations. 
This method effectively enhances the ability to model dynamics across time and 
addresses the non-stationary issue of long-term motion dynamics without 
significantly increasing the model complexity.  Differently from previous 
methods that using only feedforward connections, Shi et 
al.~\citep{shi2017learning} proposed a biologically-inspired deep 
network, called ShuttleNet1. Unlike traditional RNNs, all processors inside 
ShuttleNet are connected in a loop to mimic the human brain’s feedforward and 
feedback connections. In this manner, the processors are shared
across multiple pathways in the loop connection. 
Attention mechanism is then employed to select the best information flow 
pathway.

\begin{figure*}[t]
\begin{center}
{\includegraphics[height = 60mm, width = 175mm]{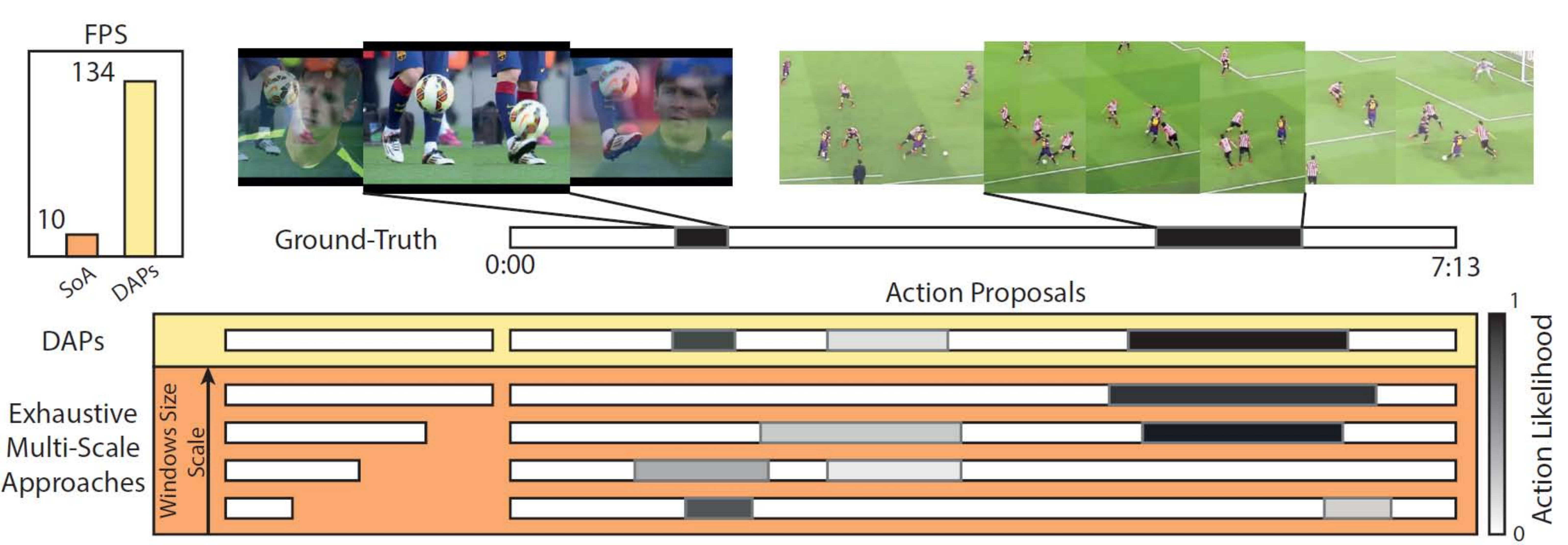}}
\end{center}
\caption{The Deep Action Proposals algorithm can localize segments of varied 
duration around actions occurring along a video without exhaustively exploring 
multiple temporal scales. Figure from~\citep{escorcia2016daps}.}
\label{DAPs}
\end{figure*}

\begin{figure}[t]
\begin{center}
{\includegraphics[height = 54mm, width = 75mm]{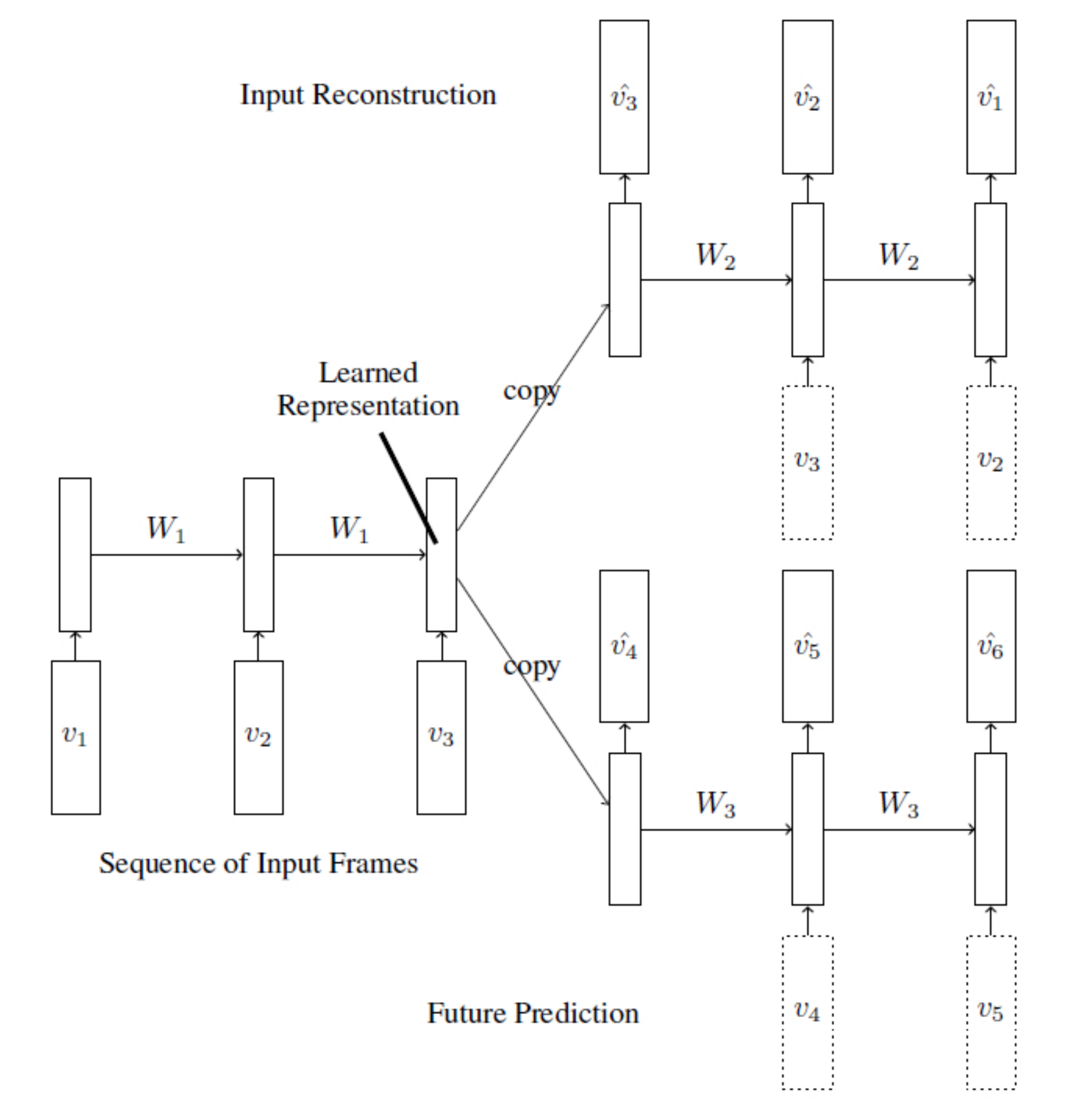}}
\end{center}
\caption{The LSTM autoencoder model and LSTM future predictor model. Figure 
from~\citep{srivastava2015unsupervised}.}
\label{unLSTM}
\end{figure}

\subsubsection{Other-architecture-based Approach} Besides the commonly used 
CNN- and RNN-based methods for motion recognition from RGB modality, there are 
several other structures that have been adopted for this task. 
Jhuang et al.~\citep{jhuang2007biologically} used a  feedforward hierarchical 
template matching architecture for action recognition with pre-defined 
spatio-temporal filters in the first layer. In his thesis, 
Chen~\citep{chen2010deep} adopted the 
convolutional RBM (CRBM) as the basic processing unit and proposed the 
so-called space-time Deep Belief Network (ST-DBN) that alternates the 
aggregation of spatial and temporal information so that higher layers capture 
longer range statistical dependencies in both space and time. 
Taylor et al.~\citep{taylor2010convolutional} extended the Gated RBM 
(GRBM)~\citep{memisevic2007unsupervised} to convolutional GRBM (convGRBM) that 
shares weights at all locations in an image and inference is performed through 
convolution.
Le et al.~\citep{le2011learning} presented an extension of the independent 
subspace analysis algorithm~\citep{theis2007towards} to learn invariant 
spatio-temporal features from unlabeled video data. They scale up the original 
ISA to larger input data by employing two important ideas from convolutional 
neural networks: convolution and stacking. This convolutional
stacking idea enables the algorithm to learn a hierarchical representation of 
the data suitable for recognition. Yan et al.~\citep{yan2014modeling} proposed 
\textit{Dynencoder}, a three layer auto-encoder, to capture video dynamics. 
Dynencoder is shown to be successful in synthesizing dynamic textures, and one 
can think of a Dynencoder as a compact way of representing the spatio-temporal 
information of a video. Similarly, Srivastava 
et al.~\citep{srivastava2015unsupervised} 
introduced a LSTM autoencoder model. The LSTM autoencoder model consists of two 
RNNs, namely, the encoder LSTM and the decoder LSTM. The encoder LSTM accepts a 
sequence as input and learns the corresponding compact representation. The 
states of the encoder LSTM contain the appearance and dynamics of the sequence. 
The decoder LSTM receives the learned representation to reconstruct the input 
sequence. Inspired by the Generative Adversarial Networks 
(GAN)~\citep{goodfellow2014generative}, Mathieu et al.~\citep{mathieu2015deep} 
adopted the adversarial mechanism to train a multi-scale convolutional network 
to generate future frames given an input sequence.  To deal with the inherently 
blurry predictions obtained from the standard Mean Squared Error (MSE) loss 
function, they proposed three different and complementary feature learning 
strategies: a multi-scale architecture, an adversarial training method, and an 
image gradient difference loss function.


\subsection{Continuous/Online Motion Recognition}
Most of the action recognition methods reviewed above rely heavily on
segmented videos for model training. However, it is very expensive and 
time-consuming to acquire a large-scale trimmed video dataset.  
The availability of untrimmed video datasets (e.g.~\citep{caba2015activitynet, 
yeung2017every, DeGeest2016, wanchalearn,liu2017pku}) have encouraged research 
and challenges/contests in motion recognition in this domain.
\subsubsection{CNN-based Approach} Inspired by the success of 
region-proposal-based object detection using 
R-CNN~\citep{girshick14CVPR,girshick2015fast,ren2015faster}, several 
proposal-based action recognition methods from untrimmed video are proposed. 
These methods first generate a reduced number of candidate temporal 
windows, and then an action classifier discriminates each proposal independently 
into one of the actions of interest. For instance, based on the two-stream 
concept~\citep{simonyan2014two}, \citep{gkioxari2015finding} 
classified frame-based region proposals of interest using static and motion 
cues. The regions are then linked across frames based on the predictions 
and their spatial overlap; thus producing action tubes respectively for each 
action and video. Weinzaepfel et al.~\citep{weinzaepfel2015learning} also 
started from the frame-level proposals, selected the highest scoring ones, 
tracked them throughout the video, and adopted a multi-scale sliding window 
approach over tracks to detect the temporal content of an action. 
Shou et al.~\citep{shou2016temporal} proposed a multi-stage segment-based 3D CNNs 
to generate candidate segments, 
that are used to recognize actions and localize temporal boundaries. 
Peng and Schimd~\citep{peng2016multi} generated rich proposals from both RGB and 
optical flow data by using region proposal networks for frame-level action 
detection, and stacked optical flows to enhance  the discriminative power of 
motion R-CNN. Wang et al.~\citep{wang2017untrimmednets} proposed an UntrimmedNet 
to generate clip proposals that may contain action instances for untrimmed 
action recognition. Based on these clip-level representations, the 
classification module aims to predict the scores for each clip proposal and the 
selection module tries to select or rank those clip proposals.  
 Similarly in the same direction, Zhao et 
al.~\citep{zhao2017temporal} adopted explicit structural modeling in the temporal 
dimension. In their model, each complete activity instance is considered
as a composition of three major stages, namely \textit{starting}, 
\textit{course}, 
and \textit{ending}, and they introduced structured temporal pyramid pooling to 
produce a global representation of
the entire proposal. Differently from previous methods, Zhu et 
al.~\citep{zhu2017tornado} proposed a framework that integrates the 
complementary spatial and temporal information into an end-to-end trainable 
system for video action proposal, and a novel and efficient path trimming method 
is proposed to handle untrimmed video by examining actionness and background 
score pattern without using extra detectors.  To generalize R-CNN 
from 2D to 3D, Hou et al.~\citep{hou2017tube} proposed an end-to-end 3D CNN-based 
approach for action detection in videos. A Tube Proposal Network was introduced 
to leverage skip pooling in temporal domain to preserve temporal information for 
action localization in 3D volumes, and Tube-of-Interest pooling layer was 
proposed to effectively alleviate the problem with variable spatial and temporal 
sizes of tube proposals. Saha et al.~\citep{saha2017amtnet} proposed a deep net 
framework capable of regressing and classifying 3D region proposals spanning 
two successive video frames. The core of the framework is an evolution of 
classical region proposal networks (RPNs) to 3D RPNs. Similarly, Kalogeiton et 
al.~\citep{kalogeiton2017action} extended the Single Shot MultiBox Detector 
(SSD)~\citep{liu2016ssd} framework from 2D to 3D by proposing an Action Tubelet 
detector. In order to quickly  and accurately generate temporal action 
proposals, Gao et al.~\citep{gao2017turn} proposed a Temporal Unit Regression 
Network (TURN) model, that jointly predicts action proposals and refines the 
temporal boundaries by temporal coordinate regression using CNN.  Similarly, 
Singh et al.~\citep{singhonline} designed an efficient online algorithm to 
incrementally construct and label 'action
tubes' from the SSD frame level detections, making it the first real-time 
(up to 40fps) system able to perform online S/T action localisation
on the untrimmed videos. Besides the 
proposal-based methods discussed above, Lea et al.~\citep{lea2016temporal} 
introduced a new class of temporal models, called Temporal Convolutional 
Networks (TCNs), that use a hierarchy of temporal convolutions to perform 
fine-grained action segmentation or detection.

\begin{figure*}[htp]
\begin{center}
{\includegraphics[height = 50mm, width = 175mm]{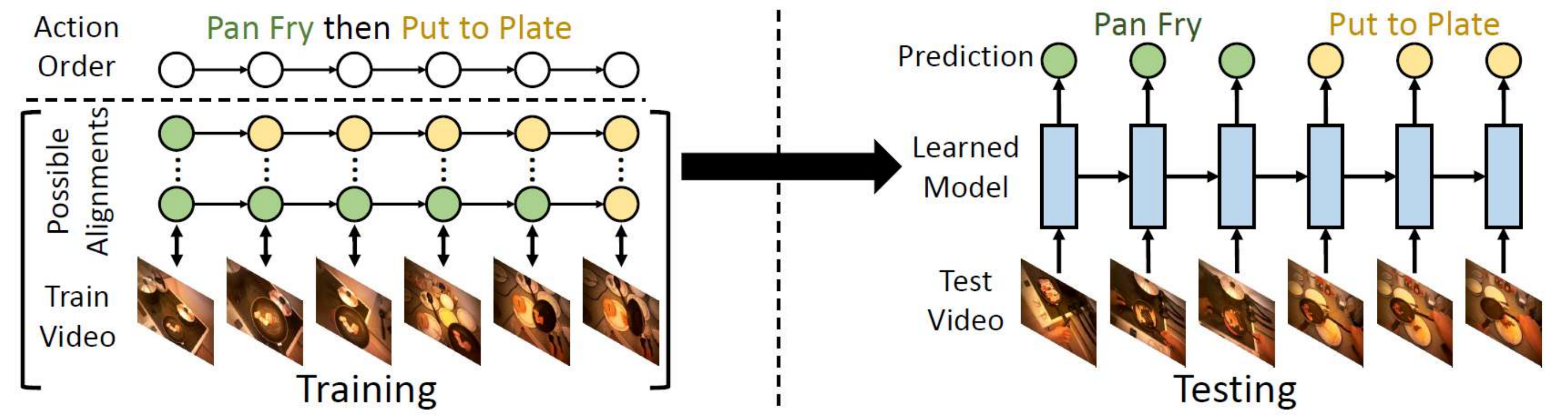}}
\end{center}
\caption{The problem of weakly supervised 
action labelling is tackled where only the order of the occurring actions in 
given during training. The temporal model is trained by maximizing the 
probability of all possible frame-to-label alignments. At testing time, no 
annotation is given. The learned model encodes the temporal structure of videos 
and could predict the actions without further information.  Figure 
from~\citep{huang2016connectionist}.}
\label{ECTC}
\end{figure*}
%

\subsubsection{RNN-based Approach} Apart from the proposal-based methods that 
use CNN, there are several proposal-based methods using RNN for temporal 
modeling. Escorcia et al.~\citep{escorcia2016daps} introduced the Deep Action 
Proposals (DAPs)  that generate temporal action proposals from long untrimmed 
videos for action detection and classification (see Figure.~\ref{DAPs}). They 
adopted C3D network~\citep{tran2015learning} for visual encoder and LSTM for 
sequence encoder. However, all of these methods generated proposals by a sliding 
window approach, dividing the video into short overlapping temporal window, 
which is computationally expensive. To reduce the number of proposals, 
Buch et al.~\citep{sst2017} proposed a single-stream temporal action proposal 
generation method that does not the need to divide input into short overlapping 
clips or temporal windows for batch processing. 

Besides the proposal-based methods, there are several methods that are 
proposal-free. Yeung et al.~\citep{yeung2016end} proposed an end-to-end training 
model that is formulated as a recurrent neural network-based agent. This agent 
learns a policy for sequentially forming and refining hypotheses about action 
instances based on the intuition that the process of detecting actions is 
naturally one of observation and refinement. They adopted two networks namely, 
observation network and recurrent network, for this purpose. 
Singh et al.~\citep{singh2016multi} presented a multi-stream bi-directional 
recurrent neural network for fine-grained action detection. They adopted a 
tracking algorithm to locate a bounding box around the person and trained two 
streams on motion and appearance cropped to the tracked bounding box. The video 
sequence was split into fixed long chunks for the input of two-stream networks, 
and  bi-directional LSTM was used to model long-term temporal dynamics within 
and between actions. 
Ma et al.~\citep{ma2016learning} introduced a novel ranking loss within the RNN 
objective so that the trained model better captures progression of activities. 
The ranking loss constrains the detection score of the correct category to be 
monotonically  non-decreasing as the activity progresses. The same time, the 
detection score margin between the correct activity category and all other 
categories is monotonically non-decreasing. Huang 
et al.~\citep{huang2016connectionist} 
proposed a weakly-supervised framework for action labeling in video ( see 
Figure~\ref{ECTC}), where only the order of occurring actions is required 
during 
training. They proposed an Extended Connectionist Temporal Classification (ECTC) 
framework to efficiently evaluate all possible alignments between the input and 
label sequences via dynamic programming and explicitly enforce their consistency 
with frame-to-frame visual similarities.  Taking inspiration from classical 
 linear dynamic systems theory for modeling time 
series, Dave et al.~\citep{dave2017predictive} derived a series of recurrent 
neural networks 
that sequentially make top-down predictions about the future and then correct 
those predictions with bottom-up observations. Their predictive-corrective 
architecture allows the incorporation of insights from time-series
analysis: adaptively focus computation on ``surprising" frames where predictions 
require large corrections; simplify learning in that only ``residual-like" 
corrective terms need to be learned over time and  naturally decorrelate an 
input stream in a hierarchical fashion, producing a more reliable signal for 
learning at each layer of a network.

\section{Depth-based Motion Recognition with Deep Learning}\label{depth}
Compared with RGB videos, the depth modality is insensitive to illumination 
variations, invariant to color and texture changes, reliable for estimating body 
silhouette and skeleton, and provides rich 3D structural information of the 
scene. However, there are only few published results on depth based action 
recognition using deep learning methods. Two reasons can be adduced for this 
situation. First, the absence of 
color and texture in depth maps weakens the discriminative representation power 
of CNN models~\citep{liu20163d}. Second, existing depth data is relatively 
small-scale. The conventional pipelines are purely data-driven and learn 
representation directly from the pixels. Such model is likely to be at risk 
of overfitting when the network is optimized on limited training data. 
Currently, there are only CNN-based methods for depth-based motion recognition.
\subsection{Segmented Motion Recognition}
\subsubsection{CNN-based Approach} 
Wang et al.~\citep{pichao2015,pichaoTHMS} proposed a method called Weighted Hierarchical Depth Motion Maps 
(WHDMM) + 3ConvNet, for human action recognition from depth maps on small 
training datasets. Three strategies were developed to leverage the 
capability of ConvNets in mining discriminative features for recognition. 
Firstly, different viewpoints are mimicked by rotating the 3D points of the 
captured depth maps. This not only auguments the data, but also makes the 
trained ConvNets view-tolerant. Secondly, WHDMMs at several temporal scales were 
constructed to encode the spatio-temporal motion patterns of 
actions into 2D spatial structures. The 2D spatial structures are further 
enhanced for recognition by converting the WHDMMs into pseudo-color images. 
Lastly, the three ConvNets were initialized with the models obtained from 
ImageNet and fine-tuned independently on the color-coded WHDMMs constructed in 
three orthogonal planes. 
 Inspired by the 
promising results achieved by rank pooling method~\citep{bilen2016dynamic} on 
RGB data, Wang et al.~\citep{wang2016large} encoded the depth map sequences into 
three kinds of dynamic images with rank pooling: Dynamic Depth Images (DDI), 
Dynamic Depth Normal Images (DDNI) and Dynamic Depth Motion Normal Images 
(DDMNI). These three representations capture the posture and motion information 
from three different levels for gesture recognition. Specifically, DDI exploits the dynamics of postures over time and DDNI and DDMNI exploit the 3D structural information captured by depth maps.  Wang et al.~\citep{pichaoTMM} replaced the bidirectional rank pooling in the method of \cite{wang2016large}  with {\em hierarchical} and {\em bidirectional} rank pooling to capture both high order and non-linear dynamics effectively for both gesture and action recognition.   Recently, Wang et al.~\citep{wang2017structured} proposed to represent a depth map sequence into three pairs of structured dynamic images at body, part and joint levels respectively through bidirectional rank pooling. Different from previous works that applied one ConvNet for each part/joint separately, one pair of structured dynamic images is constructed from depth maps at each granularity level and serves as the input of a ConvNet. The structured dynamic image not only preserves the spatial-temporal information but also enhances the structure information across both body parts/joints and at different temporal scales. In addition, it requires low computational cost and memory to construct. This new representation, referred to as  Spatially Structured Dynamic Depth Images (S$^{2}$DDI), aggregates from global to fine-grained motion and structure information in a depth sequence, and enables us to fine-tune the existing ConvNet models trained on image data for classification of depth sequences, without a need for training the models afresh. Similarly, Hou et al.~\citep{hou2017spatially} extended S$^{2}$DDI to Spatially and Temporally Structured Dynamic Depth Images (STSDDI), where a hierarchical bidirectional rank pooling method was adopted to exploit the spatio-temporal-structural information contained in the depth sequence and it is applied to interactions of two subjects.
 Differently from the above texture image encoding method, Rahmani et al.~\citep{rahmani20163d} proposed a 
cross-view action recognition based on depth sequence. Their method comprises 
two steps: 
(i) learning a general view-invariant human pose model from synthetic depth 
images and, (ii) modeling the temporal action variations. To enlarge the 
training data for CNN, they generated the training data synthetically by 
fitting realistic synthetic 3D human models to real mocap data and then 
rendering each pose from a large number of viewpoints. For spatio-temporal 
representation, they used group sparse Fourier Temporal Pyramid which  
encodes the action-specific discriminative output features of the proposed 
human pose model.

\begin{figure*}[t]
\begin{center}
{\includegraphics[height = 80mm, width = 175mm]{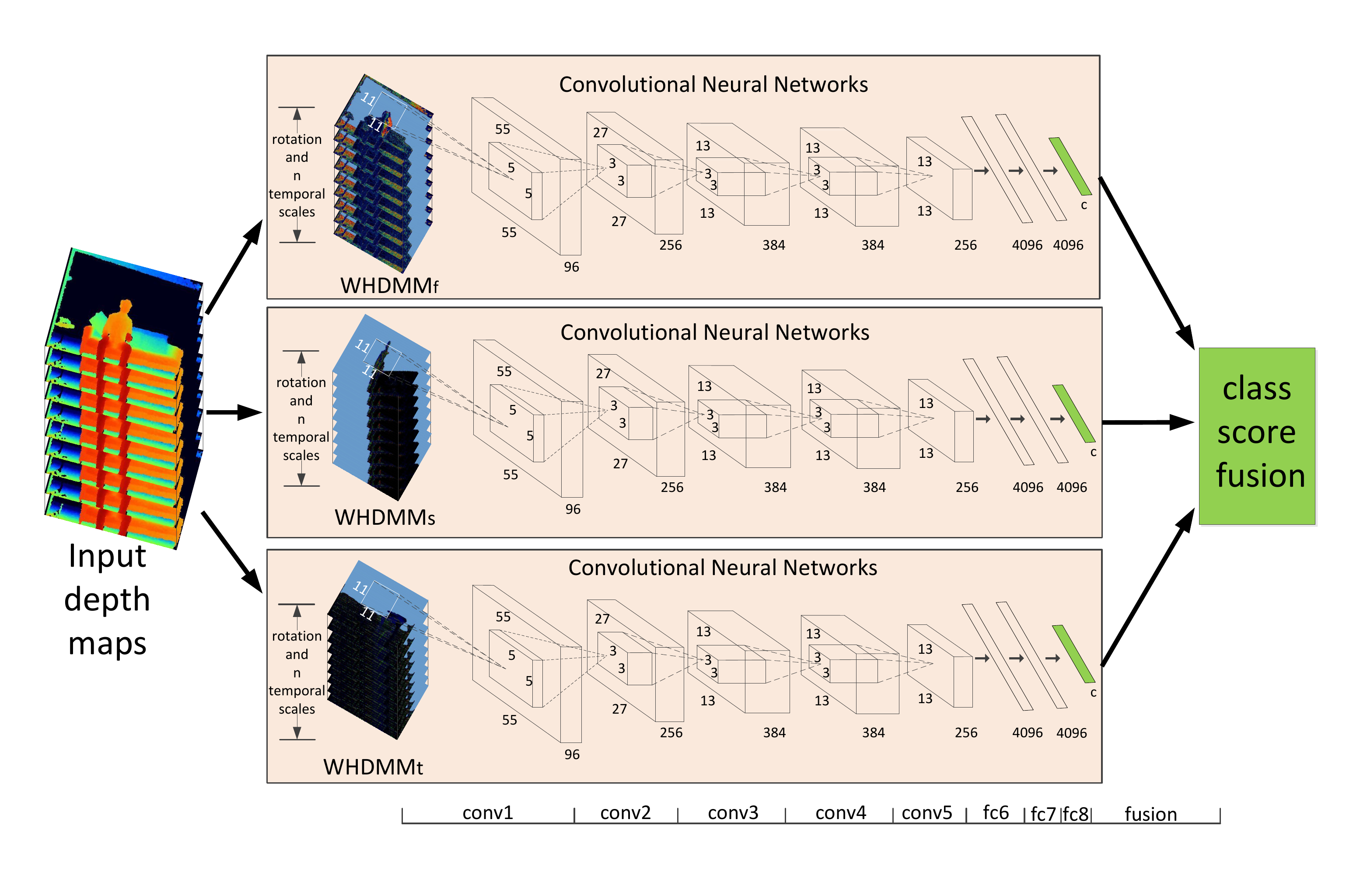}}
\end{center}
\caption{Depth map sequences are encoded into texture color images by 
using the concepts of Depth Motion Maps (DMM)~\citep{Yang2012a} and 
pseudo-coloring, and at the same time enlarged the training data by scene 
rotation on the 3D point cloud. Three channel of CNN are adopted for feature 
extraction and classification. Figure from~\citep{pichaoTHMS}.}
\label{WHDMM}
\end{figure*}

\subsection{Continuous/Online Motion Recognition}
\subsubsection{CNN-based Approach} For continuous gesture recognition, Wang et 
al.~\citep{pichaoicprwa} first segmented the continuous depth sequence 
into segmented sequences using quantity of movement 
(QOM)~\citep{jiang2015multi}, and then adopted improved DMM (IDMM) to encode the 
dynamics of depth sequences into texture images for large-scale continuous 
gesture recognition. To improve the encoding quality of depth sequences,  Wang et al.~\citep{pichaoTMM} proposed three simple, compact yet effective representations of depth sequences, referred to respectively as Dynamic Depth Images (DDI), Dynamic Depth Normal Images (DDNI) and Dynamic Depth Motion Normal Images (DDMNI), for continuous action recognition. These dynamic images are constructed from a segmented sequence of depth maps using hierarchical bidirectional rank pooling to effectively capture the spatial-temporal information. Specifically, DDI exploits the dynamics of postures over time while DDNI and DDMNI extract the 3D structural information captured by depth maps. The image-based representations enable us to fine-tune the existing ConvNet models trained on image data without training a large number of parameters from scratch.

\section{Skeleton-based Motion Recognition with Deep Learning}
\label{skeleton}
Differently from RGB and depth, skeleton data contains the positions of human joints, which can be considered relatively high-level features for motion recognition. 
There are two common ways to estimate skeletons, one is to use MOCAP systems and the other is to estimate skeletons directly from depth maps or RGB images/video. Skeletons from MOCAP systems are often robust to scale and illumination changes and can be invariant to viewpoints as well as human body rotation and motion speed; Skeletons estimated from depth maps or RGB images/video are prone to errors caused by a number of factors including viewpoints and occlusion since both factors can lead to significant different appearance of same actions.
Currently, there are mainly three approaches to skeleton-based motion 
recognition using deep learning: 
(i) RNN-based, (ii) CNN-based and other-architecture-based approaches for 
segmented motion 
recognition and, (iii) RNN-based approaches for continuous/online motion 
recognition.

\subsection{Segmented Motion Recognition}
 \begin{figure}[t]
\begin{center}
{\includegraphics[height = 50mm, width = 85mm]{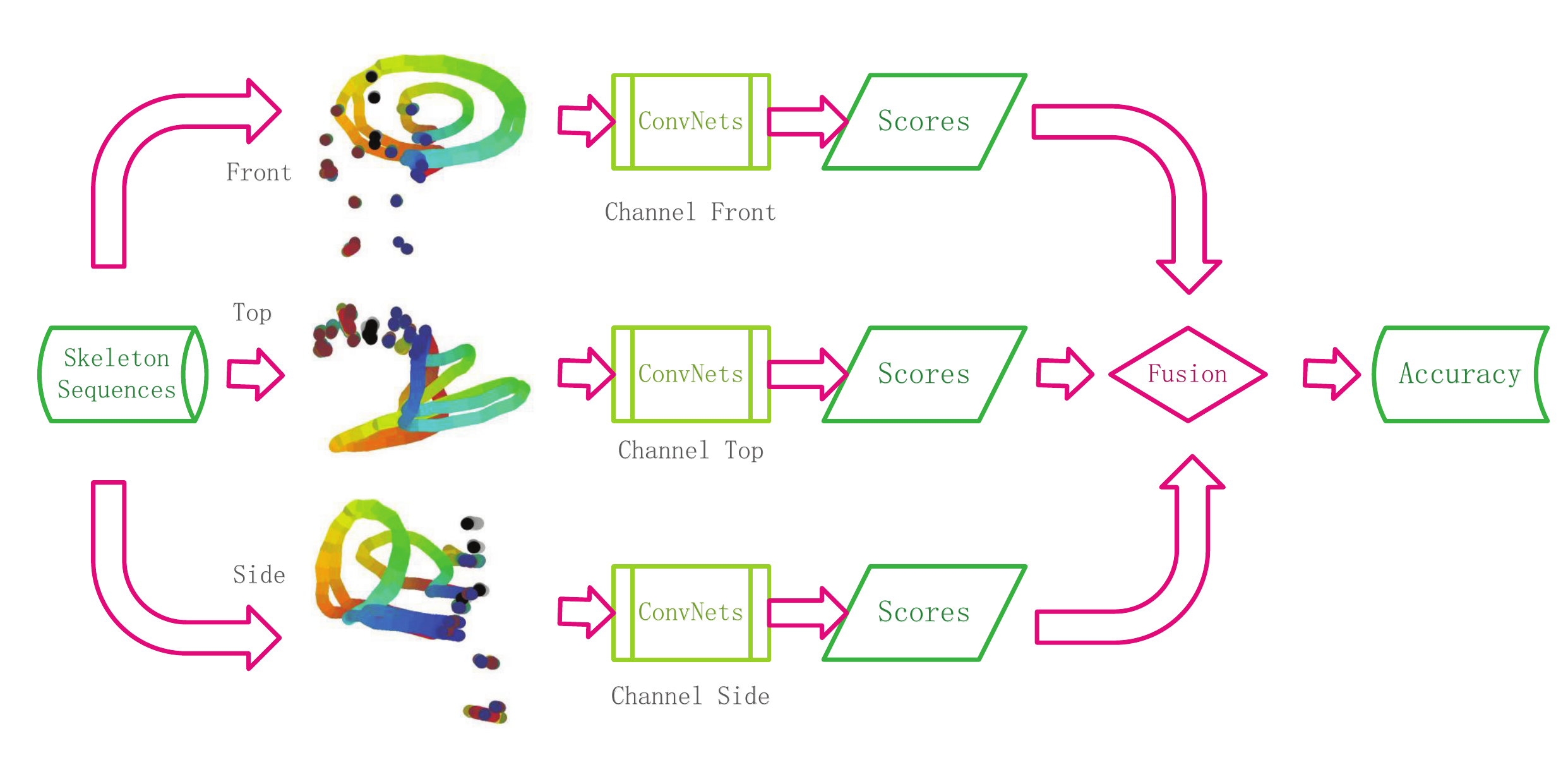}}
\end{center}
\caption{The JTM framework for skeleton-based motion recognition with CNN. 
Figure from~\citep{pichao2016}.}
\label{JTM}
\end{figure}
\subsubsection{CNN-based Approach}
The main step in this approach is to convert the skeleton sequences 
into images where the spatio-temporal information is reflected in the image 
properties including color and texture. Duet al.~\citep{du2015skeleton} 
represented a skeleton sequence as a matrix by concatenating the joint 
coordinates at each instant and arranging the vector representations in a 
chronological order. The matrix is then quantified into an image and normalized 
to handle the variable-length problem. The final image is fed into a CNN model 
for feature extraction and recognition. Wang et al.~\citep{pichao2016} proposed 
to encode spatio-temporal information contained in the skeleton sequence into 
multiple texture images, namely, Joint Trajectory Maps (JTM), by mapping the 
trajectories into HSV (hue, saturation, value) space. Pre-trained models over 
Imagenet is adopted for fine-tuning over the JTMs to extract features and 
recognize actions. Similarly, Hou et al.~\citep{pichaocsvt2016} drew the 
skeleton joints with a specific pen to three orthogonal canvases, and encodes 
the dynamic information in the skeleton sequences with color 
encoding. Li et al.~\citep{li2017joint} proposed to encode the pair-wise 
distances of skeleton joints of single or multiple subjects into texture images, 
namely, Joint Distance Maps (JDM), as the input of CNN for action recognition. 
Compared with the works reported by~\citep{pichao2016} and \citep{pichaocsvt2016}, 
JDM is less sensitive to view variations. liu et al.~\citep{liu2017enhanced} 
introduced an enhanced skeleton visualization method to represent a skeleton 
sequence as a series of visual and motion enhanced color images. They proposed 
a sequence-based view invariant transform to deal with the view variation 
problem, and multi-stream CNN fusion method is adopted to conduct recognition.
Ke et al.~\citep{ke2017skeletonnet} designed vector-based features 
for each body part of human skeleton sequences, which are translation, scale and 
rotation invariant, and transformed the features into images to feed into CNN 
for learning high level and discriminative representation. In another effort,  
Ke et al.~\citep{ke2017new} represented the sequence as a clip with several 
gray images for each channel of the 3D coordinates, which reflects multiple 
spatial structural information of the joints. The images are fed to a deep CNN 
to learn high-level features, and the CNN features of all the three clips at 
the same time-step are concatenated in a feature vector. Each feature vector 
represents the temporal information of the entire skeleton sequence and one 
particular spatial relationship of the joints. A Multi-Task Learning Network 
(MTLN) is adopted to jointly process the feature vectors of all time-steps in 
parallel for action recognition. Kim and Reiter~\citep{kim2017interpretable} 
approached the problem differently and proposed to use the Temporal 
Convolutional Neural Networks (TCN)~\citep{lea2016temporal} for skeleton based 
action recognition. They re-designed the original TCN into Res-TCN by factoring 
out the deeper layers into additive residual terms that yields both 
interpretable hidden representations and model parameters.

 \begin{figure}[t]
\begin{center}
{\includegraphics[height = 50mm, width = 85mm]{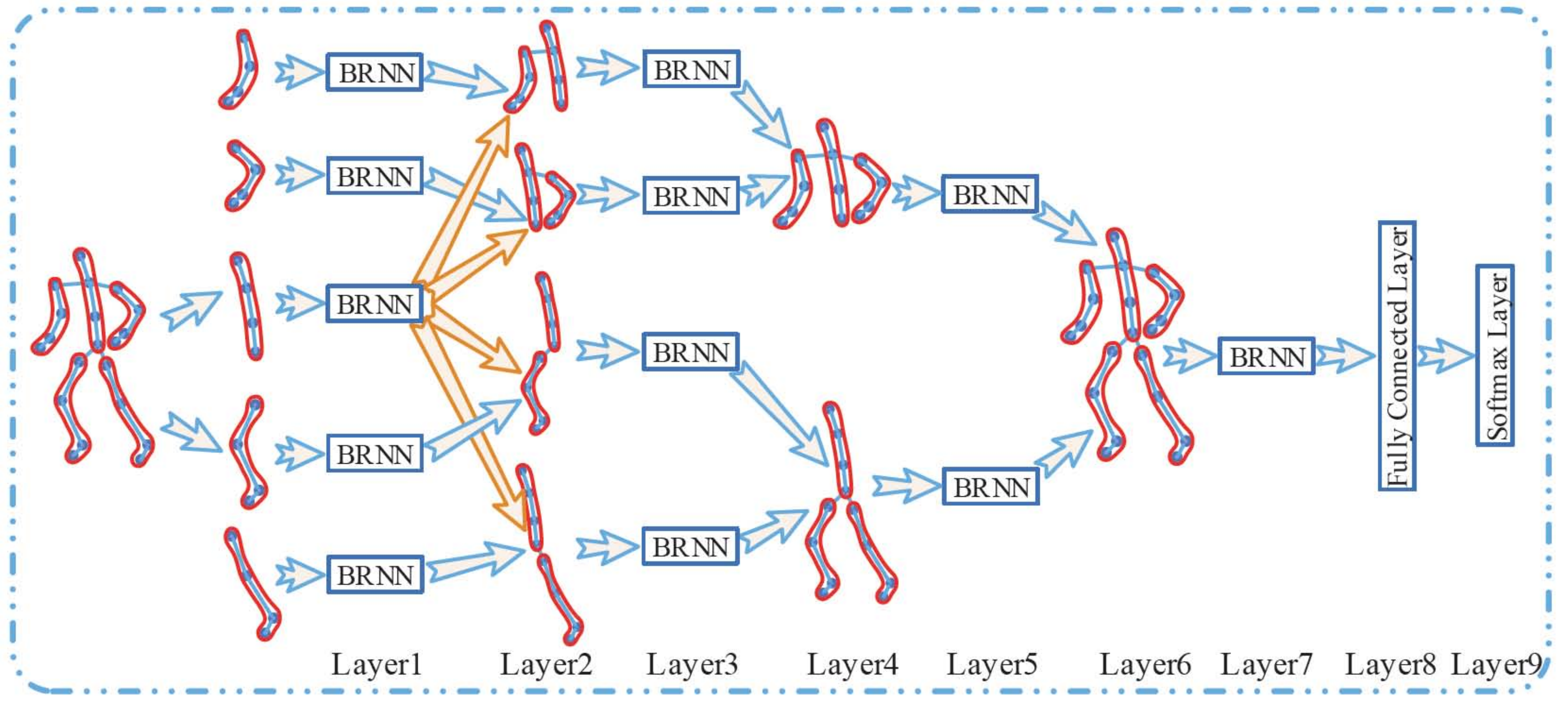}}
\end{center}
\caption{The LSTM autoencoder model and LSTM future predictor model. Figure 
from~\citep{du2015hierarchical}.}
\label{HRNN}
\end{figure}

\subsubsection{RNN-based Approach}
In this class of approaches, skeleton features are input to an RNN 
in order to exploit the temporal evolution. For 
instance, in a series or works Du 
et al.~\citep{du2015hierarchical,du2016representation} divided the whole 
skeleton sequence into five parts according to the human physical structure, 
and separately fed them into five bidirectional RNNs/LSTMs. As the number of 
layers increases, the representations extracted by the subnets are 
hierarchically fused to build a higher-level representation. The process is 
illustrated in Fig.~\ref{HRNN}. This method explicitly encodes the 
spatio-temporal-structural information into high level representation. Veeriah 
et al.~\citep{veeriah2015differential} proposed a differential gating scheme for 
the LSTM neural network, which emphasizes the change in information gain 
caused by the salient motions between the successive frames. This work is one 
of the first aimed at demonstrating the potential of learning complex 
time-series representations via high-order derivatives of states. 
Zhu et al.~\citep{zhu2015co} designed two types of regularizations to learn 
effective features and motion dynamics. In the fully connected layers, they 
introduced regularization to drive the model to learn co-occurrence features of 
the joints at different layers. Furthermore, they derived a new dropout and 
apply it to the LSTM neurons in the last LSTM layer, which helps the network to 
learn complex motion dynamics. Instead of keeping a long-term memory of the 
entire body's motion in the cell, Shahroudy et al.~\citep{shahroudy2016ntu} 
proposed a part-aware LSTM human action learning model (P-LSTM) wherein memory 
is split across part-based cells. It is argued that keeping the 
context of each body part independent and representing the output of the P-LSTM 
unit as a combination of independent body part context information is more 
efficient. Previous RNN-based 3D-action recognition methods have adopted RNN to 
model the long-term contextual information in the temporal domain for  
motion-based dynamics representation. However, there is also strong dependency 
between joints in the spatial domain. In addition the spatial configuration of 
joints in video frames can be highly discriminative for 3D-action recognition 
task. To exploit this dependency, Liu et al.\citep{liu2016spatio} proposed 
a spatio-temporal LSTM (ST-LSTM) network which extends the traditional 
LSTM-based learning to  both temporal and spatial domains. Rather than 
concatenate the joint-based input features, ST-LSTM explicitly models
the dependencies between the joints and applies recurrent analysis over spatial
and temporal domains concurrently. Besides, they introduced a trust gate 
mechanism to make LSTM robust to noisy input data. Song et 
al.~\citep{song2017end} proposed 
a spatio-temporal attention model with LSTM to automatically mine the 
discriminative joints and learn the respective and different attentions of each 
frame along the temporal axis. Similarly, Liu et al.~\citep{junliu2017} proposed 
a Global Context-Aware Attention LSTM (GCA-LSTM) to selectively focus on the 
informative joints in the action sequence with the assistance of global context 
information. Differently from previous works that adopted the coordinates of 
joints as input, Zhang et al.~\citep{zhanggeometric} investigated a set of 
simple geometric features of skeleton using 3-layer LSTM framework, and showed 
that using joint-line distances as input requires less data for training.
Based on the notion that LSTM networks with various 
time-step sizes can model various attributes well, Lee et 
al.~\citep{lee2017ensemble}  proposed an ensemble Temporal Sliding LSTM (TS-LSTM) 
networks for skeleton-based action recognition. The proposed network is
composed of multiple parts containing short-term, medium-
term and long-term TS-LSTM networks, respectively.
Li et al.~\citep{li2017adaptive} proposed an adaptive and hierarchical framework 
for fine-grained, large-scale skeleton-based action recognition. This work was 
motivated by the need to distinguish fine-grained action classes that are 
intractable using a single network, and adaptivity to new action classes by 
 model augmentation. In the framework, multiple RNNs are 
effectively incorporated in a tree-like hierarchy to mitigate the 
discriminative challenge and thus using a divide-and-conquer strategy. To deal 
with large view variations in captured human actions, Zhang et 
al.~\citep{zhang2017view} proposed a self-regulated view adaptation scheme which 
re-positions the observation viewpoints dynamically, and integrated the proposed 
view adaptation scheme into an end-to-end LSTM network which automatically 
determines the ``best” observation viewpoints during recognition.

 \begin{figure}[t]
\begin{center}
{\includegraphics[height = 50mm, width = 85mm]{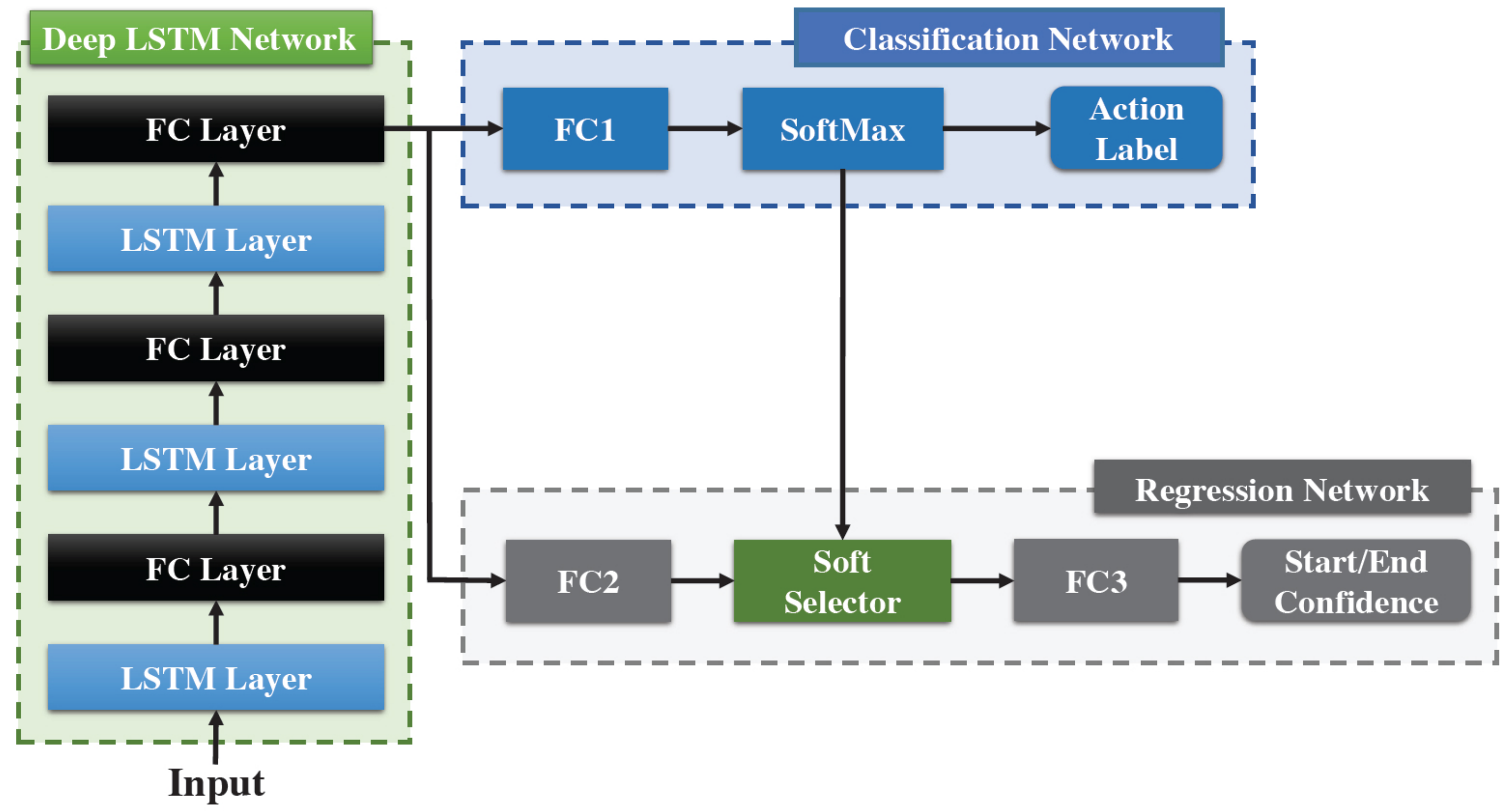}}
\end{center}
\caption{The joint classification-regression RNN framework for
online action detection and forecasting. Figure from~\citep{li2016online}.}
\label{JCRRNN}
\end{figure}

 \begin{figure*}[t]
\begin{center}
{\includegraphics[height = 65mm, width = 170mm]{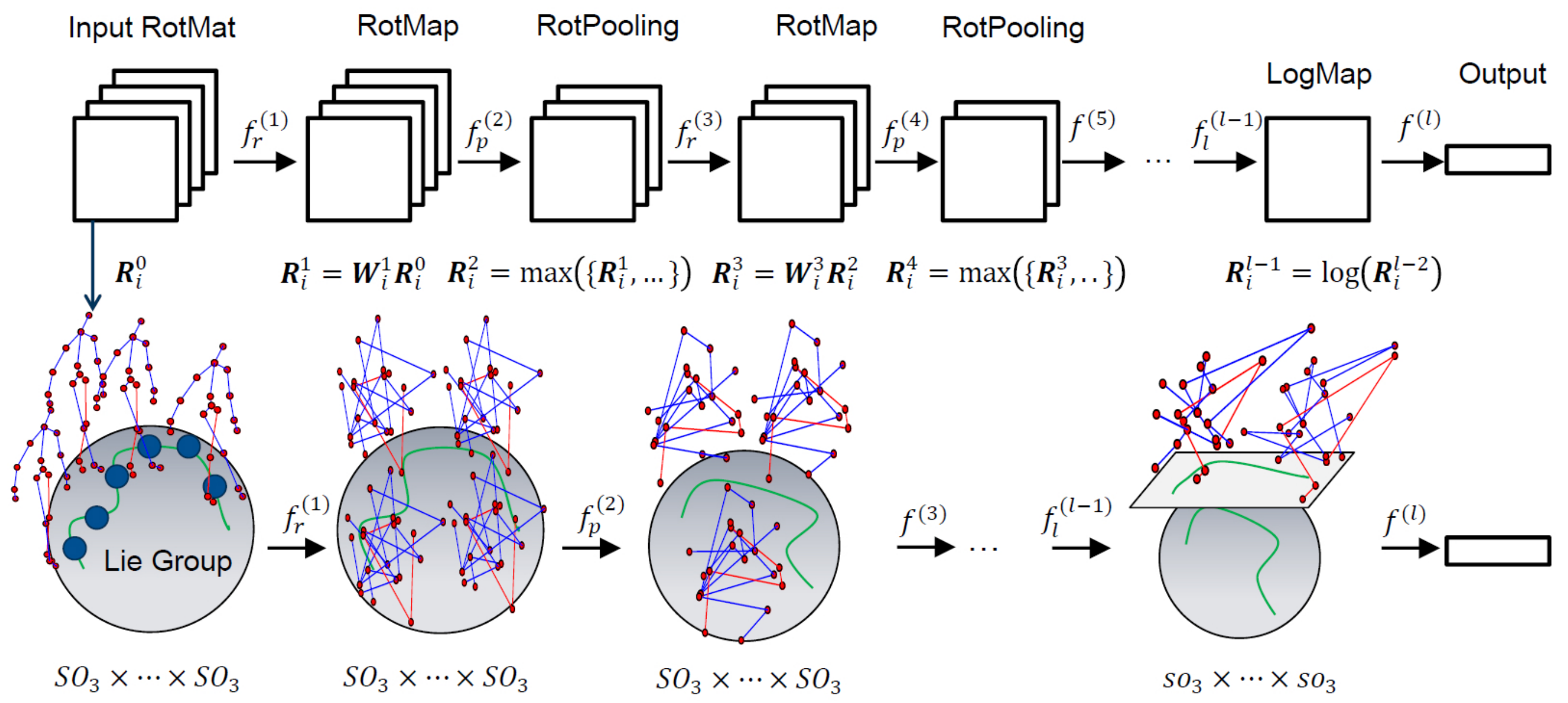}}
\end{center}
\caption{Conceptual illustration of LieNet architecture. In the network 
structure, the data space of each RotMap/RotPooling layer corresponds to a Lie 
group, while the weight spaces of the RotMap layers are Lie groups as well. 
Figure from~\citep{huang2016deep}.}
\label{LG}
\end{figure*}

\subsubsection{Other-architecture-based Approach}

Besides the RNN- and CNN-based approaches, there are several other deep 
learning-based methods. Salakhutdinov et al.~\citep{Salakhutdinov2013} proposed 
a  new compositional learning architecture that integrates deep learning
models with structured hierarchical Bayesian models. 
Specifically, this method learns a hierarchical Dirichlet process
(HDP)~\citep{teh2004sharing} prior over the activities of the 
top-level features in a deep Boltzmann machine (DBM). This compound HDP-DBM 
model learns novel concepts from very few training examples by 
learning: (i) low-level generic features, (ii) high-level features that capture 
correlations among low-level features and, (iii) a category hierarchy for 
sharing priors over the high-level features that are typical of different kinds 
of concepts. Wu and Shao~\citep{wu2014} adopted deep belief networks (DBN) to 
model the distribution of skeleton joint locations and extract high-level 
features to represent humans at each frame in 3D space. 
Ijjina et al.~\citep{ijjina2016classification} adopted stacked auto 
encoder to learn the 
underlying features of input skeleton data. Huang et al.~\citep{huang2016deep} 
incorporated the Lie group structure into a deep learning architecture to learn 
more appropriate Lie group features for skeleton based action recognition 
(see Figure.~\ref{LG}).

 \begin{figure}[t]
\begin{center}
{\includegraphics[height = 75mm, width = 85mm]{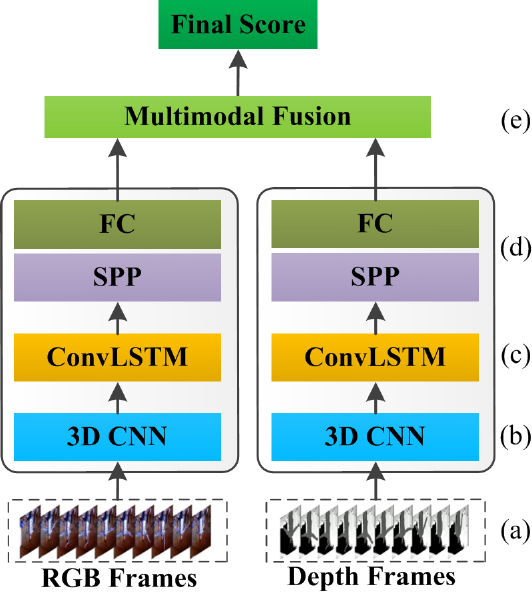}}
\end{center}
\caption{The deep architecture is composed of five 
components: (a) Input Preprocessing; (b) 3D CNN (C3D); (c) ConvLSTM; (d) Spatial 
Pyramid Pooling and Fully Connected Layers; (e) Multimodal Score Fusion. Figure 
from~\citep{zhu2017multimodal}.}
\label{ConvLSTM}
\end{figure}

\subsection{Continuous/Online Motion Recognition}

\subsubsection{RNN-based Approach}
 Differently from previous methods that recognize motion from 
 segmented skeleton sequences, Li et al.~\citep{li2016online} proposed a 
multi-task end-to-end Joint Classification-Regression Recurrent Neural Network 
to explore the action type and temporal localization information. They adopted 
LSTM to capture the complex long-range temporal dynamics, which avoids the 
typical sliding window design and thus ensures high computational efficiency. 
Furthermore, the subtask of regression optimization provides the ability to 
forecast the action prior to its occurrence. The framework is shown in 
Figure.~\ref{JCRRNN}.

\section{RGB+D-based Motion Recognition with Deep Learning}
\label{multi}
As discussed in previous sections, RGB, depth and skeleton modalities have 
their own specific properties, and how to combine the strengths of these 
modalities with deep learning approach is important.  To address this 
problem, several methods have been proposed. In general, these methods can 
be categorized as (i) CNN-based, (ii) RNN-based and 
other-architecture-based approaches for segmented motion recognition and, (iii) 
RNN-based continuous/online motion recognition.

\subsection{Segmented Motion Recognition}

\subsubsection{CNN-based Approach}
Zhu et al.~\citep{zhu2016large} fused RGB and depth in a pyramidal 
3D convolutional network based on C3D~\citep{tran2015learning} for gesture 
recognition. They designed pyramid input and pyramid fusion for each modality 
and late score fusion was adopted for final recognition. 
Duan et al.~\citep{duan2016multi} proposed a convolutional two-stream consensus 
voting network (2SCVN) which explicitly models both the short-term and 
long-term 
structure of the RGB sequences. To alleviate distractions from background, a 3D 
depth-saliency ConvNet stream (3DDSN) was aggregated in parallel to identify 
subtle motion characteristics. Later score fusion was adopted for final 
recognition. The methods described so far considered RGB and depth as 
separate channels and fused them later.  Wang et al.~\citep{Pichaocvpr2017} took 
a different approach and adopted scene flow to extract features that fused the 
RGB and depth from the onset. The new representation based on CNN and named 
Scene Flow to Action Map (SFAM) was used for motion recognition.  Different from previous methods, Wang et al.~\citep{wang2018cooperative} proposed to cooperatively train a single convolutional neural network (named c-ConvNet) on both RGB and depth features, and deeply aggregate the two kinds of features for action recognition. While the conventional ConvNet learns the deep separable features for homogeneous modality-based classification with only one softmax loss function, the c-ConvNet enhances the discriminative power of the deeply learned features and weakens the undesired modality discrepancy by jointly optimizing a ranking loss and a softmax loss for both homogeneous and heterogeneous modalities. 
Rahmani et al.~\citep{rahmani2017learning} proposed an 
end-to-end learning model for action recognition from depth and skeleton data. 
The proposed model learned to fuse features from depth and skeletal data,
capture the interactions between body-parts and/or interactions
with environmental objects, and model the temporal
structure of human actions in an end-to-end learning framework.
The proposed method was made robust to viewpoint
changes, by introducing a deep CNN which transfers visual
appearance of human body-parts acquired from different
unknown views to a view-invariant space.

\subsubsection{RNN-based Approach}
For RGB and depth fusion, Pigou et al.~\citep{pigou2015beyond} directly 
considered the depth as the fourth channel and CNN was adopted to extract 
frame-based appearance features. Temporal convolutions and RNN were combined to 
capture the temporal information.  Li et al.~\citep{li2016large} adopted 
C3D~\citep{tran2015learning} to extract features separately from RGB and depth 
modalities, and used the concatenated for SVM classifier. 
Zhu et al.~\citep{zhu2017multimodal} presented a gesture recognition method 
using C3D~\citep{tran2015learning} and convolutional LSTM 
(convLSTM)~\citep{xingjian2015convolutional} based on depth and RGB modalities 
(see Figure~\ref{ConvLSTM}). The major drawback of traditional LSTM in 
handling spatio-temporal data is its usage of full connections in 
input-to-state and state-to-state transitions in which no spatial information is 
encoded.  The ConvLSTM determines the future state of a certain cell in  the 
grid by the inputs and past states of its local neighbors. Average score fusion 
was adopted to fuse the two separate channel networks for the two modalities. 
Luo et al.~\citep{luo2017unsupervised} proposed to use a RNN-based 
encoder-decoder framework to learn a video representation by predicting a 
sequence of basic motions described as atomic 3D flows. The learned 
representation is then extracted from the generated model to recognize 
activities.

 \begin{figure}[t]
\begin{center}
{\includegraphics[height = 25mm, width = 85mm]{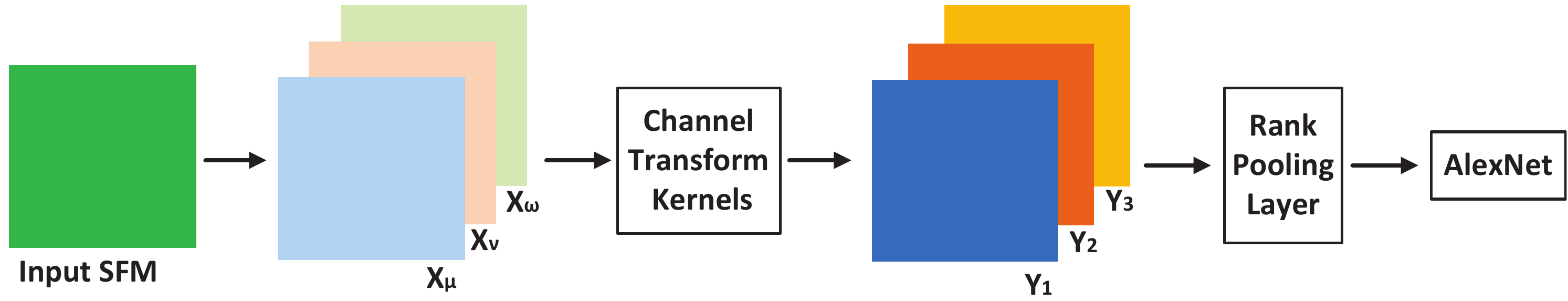}}
\end{center}
\caption{The framework of using scene flow for motion recognition. 
Scene flow vectors are first transformed into Scene Flow Maps (SFM), and then 
using Channel Transform Kernels to transform SFM into an analogous RGB space to 
take advantage of pre-train models over ImageNet. Figure 
from~\citep{Pichaocvpr2017}.}
\label{sceneflow}
\end{figure}

 \begin{figure*}[t]
\begin{center}
{\includegraphics[height = 65mm, width = 165mm]{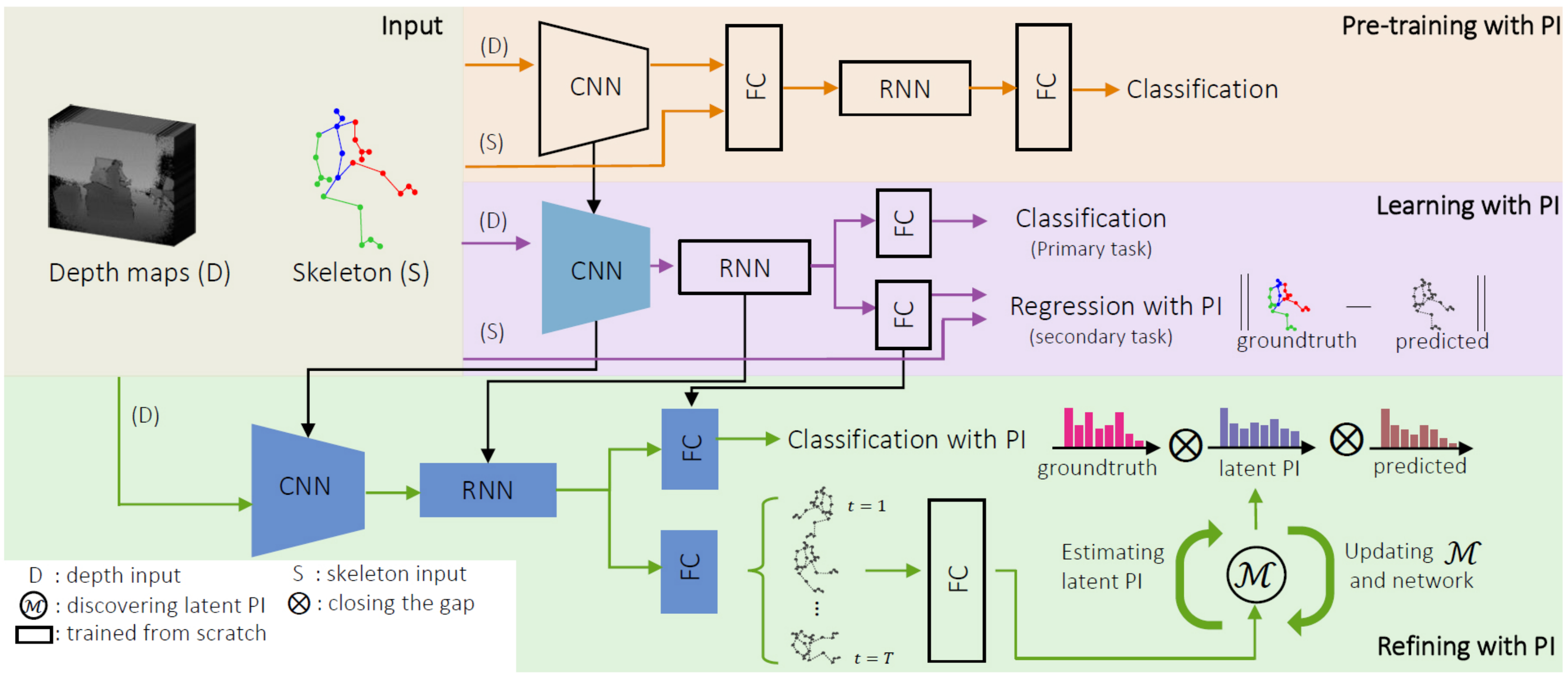}}
\end{center}
\caption{The framework of PI-based RNNs. It consists of three steps: 
(1) The pre-training step taking both depth maps and skeleton as input. An 
embedded encoder is trained in a standard CNN-RNN pipeline. (2) The trained 
encoder is used to initialize the learning step. A multi-task loss is applied to 
exploit the PI in the regression term as a secondary task. (3) Finally, 
refinement step aims to discover the latent PI by defining a bridging matrix, in 
order to maximize the effectiveness of the PI. The latent PI is utilized to 
close the gap between different information. The latent PI, bridging matrix and 
the network are optimized iteratively in an EM procedure.  Figure 
from~\citep{Shi2017}.}
\label{PIRNN}
\end{figure*}

Shi et al.~\citep{Shi2017} fused depth and skeleton in a so-called 
privileged information (PI)-based RNN (PRNN) that exploits additional 
knowledge of skeleton sequences to obtain a better estimate of network 
parameters from depth map sequences. A bridging matrix is defined to connect 
softmax classification loss and regression loss by discovering latent PI in the 
refinement step. The whole process is illustrated in Figure~\ref{PIRNN}.

For RGB and skeleton fusion, Mahassein and 
Todorovic~\citep{mahasseni2016regularizing} presented a regularization of LSTM 
learning where the output of another encoder LSTM (eLSTM) grounded on 3D 
human-skeleton training data is used as the regularization. This regularization 
rests on the hypothesis that since videos and skeleton sequences are about human 
motions their respective feature representations should be similar. The skeleton 
sequences, being view-independent and devoid of background clutter, are expected 
to facilitate capturing important motion patterns of human-body joints in 3D 
space.

\subsubsection{Other-architecture-based Approach}
Shahroudy et al.~\citep{shahroudy2017deep} extracted hand-crafted features which 
are neither independent nor fully correlated from RGB and depth, and embedded 
the input feature into a space of factorized common and modality-specific 
components. The combination of shared and specific components in input features 
can be very complex and highly non-linear. In order to disentangle them, they 
stacked layers of non-linear auto encoder-based component factorization to form 
a deep shared-specific analysis network.

In a RGB, depth and skeleton fusion method, Wu et al.~\citep{wu2016deep} 
adopted Gaussian-Bernouilli Deep Belief Network(DBN) to extract 
high-level skeletal joint features and the learned representation is used to 
estimate the emission probability needed to infer gesture sequences. A 3D 
Convolutional Neural Network (3DCNN) was used to extract features from 2D 
multiple channel inputs such as depth and RGB images stacked along the 1D 
temporal domain. In addition, intermediate and late fusion strategies were 
investigated in combination with the temporal modeling. The result of both 
mechanisms indicates that multiple-channel fusion can outperform individual 
modules.

\subsection{Continuous/Online Motion Recognition}

\subsubsection{RNN-based Approach}
Chai et al.~\citep{chai2016two} proposed to fuse RGB and depth in 
a two-stream RNN (2S-RNN) for gesture recognition. They designed a fusion layer 
for depth and RGB before the LSTM layer. \citep{molchanov2016online} 
presented an algorithm for joint segmentation 
and classification of dynamic hand gestures from continuous video streams. They 
proposed a network that employs a recurrent C3D with connectionist temporal 
classification (CTC)~\citep{graves2006connectionist}. They trained a separate 
network for each modality and averaged their scores for final recognition. 
Beside RGB and depth modalities, they also adopted stereo-IR modality in their 
work.

\section{Discussion}
\label{discuss}

We presented a comprehensive overview of RGB-D based motion recognition using 
deep learning. We defined a taxonomy covering  two groups: segmented and 
continuous/online motion recognition, with four categories in each group based 
on the adopted modalities. From the viewpoint of encoding 
spatio-temporal-structural information contained in the video sequences, CNN, 
RNN and other networks adopted for motion recognition are discussed in each 
category. In subsequent sections, the relative performance of the different 
methods on several commonly used RGB-D datasets are analysed, and from the 
comparisons we highlight some challenges. The discussion on
performance and challenges then provides a basis for outlining potential 
future research directions.

\subsection{Performance Analysis of the Current Methods}
In this section, we compare the accuracy of different methods using 
several commonly used datasets, including CMU Mocap, HDM05, MSR-Action3D, 
MSRC-12, MSRDailyActivity3D, UTKinect,  G3D, SBU Kinect Interaction, Berkeley 
MHAD, Northwestern-UCLA Multiview Action3D, ChaLearn LAP IsoGD, NTU RGB+D, 
ChaLearn2014, ChaLearn LAP ConGD, and PKU-MMD. These datasets cover motion 
capture sensor system, structured light cameras (Kinect v1) and ToF cameras 
(Kinect v2). The last three datasets are continuous datasets while the others 
are segmented datasets. 
The performance is evaluated using \textit{accuracy} for segmented motion 
recognition, and \textit{Jaccard Index} is added as another criteria for 
continuous motion recognition. The accuracy is calculated as the proportion of 
accurately labelled samples. 
The Jaccard index measures the average relative overlap between true and 
predicted sequences of frames for a given gesture/action. For a sequence $s$, let 
$G_{s,i}$ and $P_{s,i}$ be binary indicator vectors for which 1-values 
correspond to frames in which the $i^{th}$ gesture/action label is being performed. The 
Jaccard Index for the $i^{th}$ class is defined for the sequence $s$ as:

\begin{equation}
J_{s,i} = \frac{G_{s,i}\bigcap P_{s,i}}{G_{s,i}\bigcup P_{s,i}},
\end{equation}
where $G_{s,i}$ is the ground truth of the $i^{th}$ gesture/action label in 
sequence $s$, and $P_{s,i}$ is the prediction for the $i^{th}$ label in sequence 
$s$. When $G_{s,i}$ and $P_{s,i}$ are empty, $J_{(s,i)}$ is defined to be 0. 
Then for the sequence $s$ with $l_{s}$ true labels, the Jaccard Index $J_{s}$ is 
calculated as:

\begin{equation}
J_{s} = \frac{1}{l_{s}}\sum_{i=1}^{L}J_{s,i}.
\end{equation}
For all test sequences $S = {s_{1},...,s_{n}}$ with $n$ gestures/actions, the 
mean Jaccard Index $\overline{J_{S}}$ is used as the evaluation criteria and 
calculated as:
 \begin{equation}
 \overline{J_{S}} = \frac{1}{n}\sum_{j=1}^{n}J_{s_{j}}.
 \end{equation}
 
The detailed comparison of different methods is presented in following 
Table~\ref{performance}. 
From the Table we can see that there is no single approach that is able to 
produce the best performance over all datasets. Generally speaking, methods 
using multi-modal information can have better performance than their single 
modality counterpart due to the complementary properties of the three different 
modalities. On some datasets, such as NTU RGB+D dataset, current results 
suggest that CNN-based methods tend to be better than RNN-based methods. This 
is probably due to fact that CNN-based methods includes human empirical 
knowledge in the coding process, and could take advantage of pre-trained models 
over large image set, such as ImageNet. The combination of CNN and RNN seems to 
be a good choice for motion recognition, for instance, the 
C3D+ConvLSTM~\citep{zhu2017multimodal} method achieved promising 
results on ChaLearn LAP IsoGD dataset. For continuous motion recognition, 
RNN-based methods tend to achieve good results.

\subsection{Challenges}
The advent of low-cost RGB-D sensors that have access to extra depth and 
skeleton data, has motivated the significant development of human motion 
recognition. Promising results have been achieved with deep learning 
approaches~\citep{pichaoTHMS, zhanggeometric,liu2016spatio}, on several 
constrained simple datasets, such as MSR-Action3D, Berkeley MHAD and SBU Kinect 
Interaction. Despite this success, results are far from 
satisfactory on some large complex datasets, such as ChaLearn LAP IsoGD and NTU 
RGB+D datasets and especially the continuous/online datasets.  In fact, it 
is still very difficult to build a practical intelligent recognition system. 
Such goal poses several challenges:

\textbf{Encoding temporal information.} As discussed, there are several methods 
to encode temporal information. We can use CNN to extract frame-based features 
and then conduct temporal fusion~\citep{karpathy2014large}, or adopt 3D filter 
and 3D pooling layers to learn motion features~\citep{tran2015learning}, or use 
optical/scene flow to extract motion information~\citep{simonyan2014two, 
Pichaocvpr2017}, or encode the video into 
images~\citep{bilen2016dynamic,pichaoTHMS,pichao2016}, or use RNN/LSTM to model 
the temporal dependences~\citep{donahue2015long,du2015hierarchical,junliu2017}. 
However, all these approaches have their drawbacks. Temporal fusion method 
tends to neglect the temporal order; 3D filters and 3D pooling filters have a 
very rigid temporal structure and they only accept a predefined number of frames 
as input which is always short; optical/scene flow methods are computationally 
expensive; sequence to images methods inevitably loses temporal information 
during encoding; the weight sharing mechanism of RNN/LSTM methods make the 
sequence matching imprecise, but rather approximated, so an appropriate 
distance function must be used to predict the match probability. In fact, there 
is still no perfect method for temporal encoding, and how to model temporal 
information is a big challenge.

\textbf{Small training data.} Most of available deep learning methods rely on 
large labeled training data~\citep{karpathy2014large,tran2015learning}. 
However, in practical scenarios, obtaining large labeled training data is 
costly and laborious, even impossible, especially in medical-related 
applications. It has been shown that fine-tuning motion-based networks with 
spatial data (ImageNet) is more effective than training from 
scratch~\citep{simonyan2014two,pichao2016,bilen2016dynamic,Pichaocvpr2017}. 
Strategies for data augmentation are also commonly used~\citep{pichaoTHMS}. 
Likewise, training mechanisms to avoid overfitting and control learning rate 
have also been studied~\citep{srivastava2014dropout}. However, it is still a 
challenge to effectively train deep networks from small training data.

\textbf{Viewpoint variation and occlusion.} When skeletons are estimated from RGB images/video or depth maps, viewpoint variation may cause significantly different appearance of same actions, and occlusion would ``crash'' the skeleton data. Occlusion includes inter-occlusion caused by other 
subjects or objects, and self-occlusion created by the object/subject itself. 
Most of available datasets require subjects to perform actions in a visible and 
restricted view to avoid occlusion, and this results in limited view data 
collection and less occlusion. However, occlusion is inevitable in practical 
scenarios, especially for interactions. This makes it challenging to isolate 
individuals in overlapping area and extract features of a unique person; 
leading to the ineffectiveness of many of available 
approaches~\citep{du2015hierarchical,shahroudy2016ntu,li2017joint}. Possible  
solutions to handle viewpoint variation and occlusion include the use of 
multi-sensor 
systems~\citep{ofli2013berkeley,wang2014cross,shahroudy2016ntu,liu2017pku}. The 
multi-camera systems is able to generate multi-view data, but the drawback is 
the requirement of synchronization and feature/recognition fusion among 
different views. This usually increases processing complexity and 
computation cost. Several methods have been proposed to handle the viewpoint 
variation and occlusion. Wang et al.~\citep{pichao2015} proposed to rotate the 
depth data in 3D point clouds through different angles to deal with viewpoint 
invariance; spherical coordinate system corresponding to body center was 
developed to achieve view-independent motion recognition~\citep{huang2016deep}. 
However, these methods become less effective when occlusion occurs. How to 
effectively handle occlusion using deep learning methods is a new challenge. 

\textbf{Execution rate variation and repetition.}  The execution rate may vary 
due to the different performing styles and states of individuals. The varying 
rate results in different frames for the same motion. Repetition also bring 
about this issue. The global encoding 
methods~\citep{pichaocsvt2016,ke2017skeletonnet,liu2017enhanced} would become 
less effective due to the repetition. The commonly used methods to handle this 
problem is up/down sampling~\citep{zhu2015co,zhanggeometric,li2017joint}. 
However, sampling methods would inevitably bring redundant or loss of useful 
information. Effective handling of this problem remains a challenge.

\textbf{Cross-datasets.} Many research  works have been carried out to recognize 
human actions from RGB-D video clips. 
To learn an effective action classifier, most of the previous approaches rely 
on enough training labels.  When being required to recognize the action in a 
different dataset, these approaches have to re-train the model using new labels. 
However, labeling video sequences is a very tedious and time-consuming task, 
especially when detailed spatial locations and time durations are required. Even 
though some works have studied this 
topic~\citep{cao2010cross,Sultani,zhang2017joint}, they are all based on 
hand-crafted features, and the results are far from satisfactory due to the 
large distribution variances between different datasets, including different 
scenarios, different modalities,  different views, different persons, and even 
different actions. How to deal with cross-datasets RGB-D motion recognition is a 
big challenge.
 
 \begingroup
 \begin{table*}
 \afterpage{\onecolumn  \normalsize
 \topcaption{Performance comparison among different methods on commonly used 
 RGB-D datasets. 
 Notation: D:Depth, S:Skeleton, Acc:Accuracy, JI:Jaccard Index, cs:cross-subject 
 setting, cv:cross-view setting. Without specific notation, accuracy is used for 
 the metric.\label{performance}}
 \begin{xtabular*}{\linewidth}{c@{\extracolsep{\fill}}|c|c|c|c|c}\hline
  Dataset                              & Reference                         & Modality           & Method                     & \tabincell{c}{Fusion\\Method}       & Metric \\\hline                                
 CMU Mocap                            & \citep{zhu2015co}                & Skeleton           &\tabincell{c}{Co-occurrence+LSTM}       & None                & 81.04\%  \\
                                      & \citep{ke2017new}                & Skeleton           & \tabincell{c}{Clips+CNN+MTLN}             & None                & 88.30\%  \\\hline
 HDM05                                & \citep{huang2016deep}            & Skeleton           & \tabincell{c}{Deep Learning\\ on Lie Group} & None                & 75.78\%  \\
                                      & \citep{du2015hierarchical}       & Skeleton           & HBRNN-L                    & None                & 96.92\%  \\
                                      & \citep{zhu2015co}                & Skeleton           & \tabincell{c}{Co-occurrence+LSTM}       & None                & 97.25\%  \\\hline
 MSR-Action3D                         & \citep{liu20163d}                & Depth              & 3DCNN                      & None                & 84.07\%  \\
                                      & \citep{veeriah2015differential}  & Skeleton           & dRNN                       & None                & 92.03\%  \\
                                      & \citep{du2015hierarchical}       & Skeleton           & HBRNN-L                    & None                & 94.49\%  \\
                                      & \citep{Shi2017}                  & \tabincell{c}{Depth+\\Skeleton}     & PRNN                       & \tabincell{c}{Side\\Information}    & 94.90\%  \\
                                      & \citep{pichaoTHMS}               & Depth              & WHDMM+CNN                  & None                & 100.00\% \\
                                                                            & \citep{wang2017structured}               & Depth              & S$^{2}$DDI                  & None                & 100.00\% \\\hline
 MSRC-12                              & \citep{pichao2016}               & Skeleton           & JTM+CNN                    & None                & 93.12\%  \\
                                      & \citep{pichaocsvt2016}           & Skeleton           & SOS+CNN                    & None                & 94.27\%  \\
                                      & \citep{liu2017enhanced}          & Skeleton           & \tabincell{c}{Enhanced Visualization+CNN} & None                & 96.62\%  \\\hline
 \tabincell{c}{MSRDaily\\Activity3D}                   & \citep{pichaoTHMS}               & Depth              & WHDMM+CNN                  & None                & 85.00\%  \\
                                      & \citep{luo2017unsupervised}      & Depth              & \tabincell{c}{Unsupervised\\+ConvLSTM}      & None                & 86.90\%  \\
                                      & \citep{shahroudy2017deep}        & \tabincell{c}{RGB+\\Depth}          & DSSCA-SSLM                 & \tabincell{c}{Hierarchical\\Fusion} & 97.50\%  \\
                                      & \citep{wang2017structured}        & \tabincell{c}{Depth}          & S$^{2}$DDI                 & None & 97.50\%  \\\hline
 UTKinect                             & \citep{liu20163d}                & Depth              & 3DCNN                      & None                & 82.00\%  \\
                                      & \citep{pichaoTHMS}               & Depth              & WHDMM+CNN                  & None                & 90.91\%  \\
                                      & \citep{zhanggeometric}           & Skeleton           & JL\_d+RNN                  & None                & 95.96\%  \\
                                      & \citep{lee2017ensemble}          & Skeleton           & Ensemble TS-LSTM  & Ensemble & 96.97\% \\
                                      & \citep{liu2016spatio}            & Skeleton           & ST-LSTM+Trust Gate         & None                & 97.00\%  \\\hline
 G3D                                  & \citep{huang2016deep}            & Skeleton           & \tabincell{c}{Deep Learning\\ on Lie Group} & None                & 89.10\%  \\
                                      & \citep{pichao2016}               & Skeleton           & JTM+CNN                    & None                & 94.24\%  \\
                                      & \citep{pichaocsvt2016}           & Skeleton           & SOS+CNN                    & None                & 95.45\%  \\
                                       & \citep{wang2017structured}           & Depth           & S$^{2}$DDI                    & None                & 96.06\%  \\\hline
 \tabincell{c}{SBU Kinect\\Interaction}               & \citep{Shi2017}                  & \tabincell{c}{Depth+\\Skeleton}     & PRNN                       & \tabincell{c}{Side\\Information}    & 89.20\%  \\
                                      & \citep{zhu2015co}                & Skeleton           & \tabincell{c}{Co-occurrence\\+LSTM}       & None                & 90.41\%  \\
                                      & \citep{song2017end}              & Skeleton           & STA-LSTM                   & None                & 91.51\%  \\
                                      & \citep{liu2016spatio}            & Skeleton           & ST-LSTM+Trust Gate         & None                & 93.30\%  \\
                                      & \citep{ke2017skeletonnet}        & Skeleton           & SkeletonNet(CNN)           & None                & 93.47\%  \\
                                      & \citep{ke2017new}                & Skeleton           & Clips+CNN+MTLN             & None                & 93.57\%  \\
                                      & \citep{zhanggeometric}           & Skeleton           & JL\_d+RNN                  & None                & 99.02\%  \\\hline
 Berkeley MHAD                        & \citep{ijjina2016classification} & Skeleton           & Stacked Autoencoder        & None                & 98.03\%  \\
                                      & \citep{du2015hierarchical}       & Skeleton           & HBRNN-L                    & None                & 100.00\% \\
                                      & \citep{zhanggeometric}           & Skeleton           & JL\_d+RNN                  & None                & 100.00\% \\
                                      & \citep{du2015skeleton}           & Skeleton           & Skeleton Matrix+CNN        & None                & 100.00\% \\
                                      & \citep{liu2016spatio}            & Skeleton           & ST-LSTM+Trust Gate         & None                & 100.00\% \\\hline
 \tabincell{c}{Multiview\\Action3D} & \citep{rahmani20163d}            & Depth              & Fitting model+CNN          & None                & 92.00\%  \\
                                      & \citep{liu2017enhanced}          & Skeleton           & \tabincell{c}{Enhanced Visualization\\+CNN} & None                & 92.61\%  \\\hline
 \tabincell{c}{ChaLearn LAP\\IsoGD}                   & \citep{Pichaocvpr2017}           & \tabincell{c}{RGB+\\Depth}        & Sceneflow+CNN              & Early Fusion        & 36.27\%  \\
                                      & \citep{wang2016large}            & Depth              & DynamicImages+CNN          & None                & 39.23\%  \\
                                      & \citep{pichaoTMM}            & Depth              & DynamicMaps+CNN          & None                & 43.72\%  \\
                                      & \citep{wang2018cooperative}            & Depth+RGB              & Cooperative CNN          & None                & 44.80\%  \\
                                      & \citep{zhu2016large}             & \tabincell{c}{RGB+\\Depth}          & Pyramidal C3D              & Score Fusion        & 45.02\%  \\
                                      & \citep{duan2016multi}            & \tabincell{c}{RGB+\\Depth}          & 2SCVN-3DDSN                & Score Fusion        & 49.17\%  \\
                                      & \citep{yunanli}                  & \tabincell{c}{RGB+\\Depth}          & C3D                        & Score Fusion        & 49.20\%  \\
                                      & \citep{zhu2017multimodal}        & \tabincell{c}{RGB+\\Depth}          & C3D+ConvLSTM               & Score Fusion        & 51.02\%  \\\hline
 NTU RGB+D                            & \citep{luo2017unsupervised}      & RGB                & \tabincell{c}{Unsupervised\\+ConvLSTM}     & None                & 56.00\%(cs)  \\
                                      & \citep{huang2016deep}            & Skeleton           & \tabincell{c}{Deep Learning \\on Lie Group} & None                & \tabincell{c}{61.37\%(cs)\\66.95\%(cv)}  \\
                                      & \citep{shahroudy2016ntu}         & Skeleton           & 2Layer P-LSTM              & None                & \tabincell{c}{62.93\%(cs)\\70.27\%(cv)}  \\
                                      & \citep{luo2017unsupervised}      & Depth              & \tabincell{c}{Unsupervised\\+ConvLSTM}     & None                & 66.20\%(cs)  \\
                                      & \citep{liu2016spatio}            & Skeleton           & ST-LSTM+Trust Gate         & None                & \tabincell{c}{69.20\%(cs)\\77.70\%(cv)}  \\
                                      & \citep{zhanggeometric}           & Skeleton           & JL\_d+RNN                  & None                & \tabincell{c}{70.26\%(cs)\\82.39\%(cv)}  \\
                                      & \citep{pichao2016}               & Skeleton           & JTM+CNN                    & None                & \tabincell{c}{73.40\%(cs)\\75.20\%(cv)}  \\
                                      & \citep{song2017end}              & Skeleton           & STA-LSTM                   & None                & \tabincell{c}{73.40\%(cs)\\81.20\%(cv)}  \\
                                      & \citep{kim2017interpretable}     & Skeleton           & Res-TCN                    & None                & \tabincell{c}{74.30\%(cs)\\83.10\%(cv)}  \\
                                      & \citep{lee2017ensemble}          & Skeleton           & Ensemble TS-LSTM  & Ensemble & \tabincell{c}{74.60\%(cs)\\81.25\%(cv)} \\
                                      & \citep{shahroudy2017deep}        & \tabincell{c}{RGB+\\Depth}          & DSSCA-SSLM                 & Hierarchical Fusion & 74.86\%(cs)  \\
                                      & \citep{ke2017skeletonnet}        & Skeleton           & SkeletonNet(CNN)           & None                & \tabincell{c}{75.94\%(cs)\\81.16\%(cv)}  \\
                                      & \citep{li2017joint}              & Skeleton           & JDM+CNN                    & None                & \tabincell{c}{76.20\%(cs)\\82.30\%(cv)}  \\
                                      & \citep{ke2017new}                & Skeleton           & Clips+CNN+MTLN             & None                & \tabincell{c}{79.57\%(cs)\\84.83\%(cv)}  \\
                                      & \citep{liu2017enhanced}          & Skeleton           & \tabincell{c}{Enhanced Visualization\\+CNN} & None                & \tabincell{c}{80.03\%(cs)\\87.21\%(cv)} \\
                                       & \citep{zolfaghari2017chained}          & RGB           & \tabincell{c}{Chained Multi-stream} & Markov chain                & 80.80\%(cs) \\                   
                                      & \citep{wang2018cooperative} &  RGB+Depth&Cooperative CNN &None &\tabincell{c}{86.42\%(cs)\\89.08\%(cv)} \\ 
                                      & \citep{pichaoTMM} &  Depth&DynamicMaps+CNN &None &\tabincell{c}{87.08\%(cs)\\84.22\%(cv)} \\ \hline                                      
                                      ChaLearn2014                         & \citep{du2015skeleton}           & Skeleton           & Skeleton Matrix+CNN        & None                & 91.16\%  \\
                                      & \citep{pigou2015beyond}          & RGB                & Temp Conv+RNN,LSTM         & None                & \tabincell{c}{94.49\%\\0.906(JI)}  \\
                                      & \citep{molchanov2016online}      & \tabincell{c}{RGB+\\Depth}          & CTC+RNN                    & Score Fusion        & \tabincell{c}{98.20\%\\0.980(JI)}  \\
                                      & \citep{wu2016deep}               & \tabincell{c}{RGB+\\Depth+\\Skeleton} & DDNN                       & Late Fusion         & 0.809(JI) \\\hline
 \tabincell{c}{ChaLearn LAP\\ ConGD}                         & \citep{pichaoicprwa}           & Depth           & IDMM+CNN        & None                & 0.240(JI)  \\                                     
                                      & \citep{chai2016two}      & \tabincell{c}{RGB+\\Depth}          & RNN                    & Score Fusion        &0.266(JI) \\
                                      & \citep{Camgoz2016}          & RGB                & C3D        & None                & 0.343(JI)   \\
                                       & \citep{pichaoTMM}          & Depth                & DynamicMaps+CNN        & None                & 0.391(JI)   \\\hline                                        
 PKU-MMD                         & \citep{song2017end}           & S           & \tabincell{c}{SlidingWindow\\+STA-LSTM}        & None                & \tabincell{c}{0.427(JI)(cs)\\0.468(JI)(cv)}  \\                                     
                                      & \citep{li2016online}      & S         & JCR-RNN                    & None        &\tabincell{c}{0.479(JI)(cs)\\0.728(JI)(cv)} \\\hline
 \end{xtabular*}
 }
 \end{table*}
 \endgroup

\textbf{Online motion recognition.} Most of available methods rely on segmented 
data, and their capability for online recognition is quite limited. Even though 
continuous motion recognition is one improved version where the videos are 
untrimmed, it still assumes that all the videos are available before 
processing. Thus, proposal-based 
methods~\citep{shou2016temporal,wang2017untrimmednets} can be adopted for 
offline processing.  Differently from continuous motion recognition, online 
motion recognition aims to receive continuous streams of unprocessed visual data 
and recognize actions from an unsegmented stream of data in a continuous 
manner. 
So far two main approaches can be identified for online 
recognition, sliding window-based and  RNN-based. Sliding window-based 
methods~\citep{liu2017pku} are simple extension of segmented-based action 
recognition methods. They often consider the temporal coherence within the 
window for prediction and the window-based predictions are further fused to 
achieve online recognition. However, the performance of these methods are 
sensitive to the window size which depends on actions and is hard to set. Either 
too large or too small a window size could lead to significant drop in 
recognition. For RNN-based methods~\citep{molchanov2016online,li2016online}, 
even though promising results have been achieved, it is still far from 
satisfactory in terms of performance. How to design effective practical online 
recognition system is a big challenge.

\textbf{Action prediction.} We are faced with numerous situations in 
which we must predict what actions other people are about to do in the near 
future.  Predicting  future  actions  before  they are actually executed is a 
critical ingredient for enabling us to effectively interact  with  other  humans 
 on  a  daily  
basis~\citep{ryoo2011human,hoai2014max,lan2014hierarchical,vu2014predicting,sadegh2017encouraging}.  
There are mainly two  challenges for this task: first, we need to capture the 
subtle details inherent in human movements that may imply a future action; 
second, predictions usually
should  be  carried  out  as  quickly  as  possible  in  the  social  world,  when
limited prior observations are available. 
Predicting the action of a person before it is actually executed has a wide 
range  of  applications  in  autonomous  robots,  surveillance and health care. 
How to develop effective algorithms for action prediction is really challenging.

\subsection{Future Research Directions}
The discussion on the challenges faced by available methods allows us to 
outline several future research directions for the development of deep learning 
methods for motion recognition. While the list is not exhaustive, they point at 
research activities that may advance the field.

\textbf{Hybrid networks.} Most of previous methods adopted one type of neural 
networks for motion recognition. As discussed, there is no perfect solution for 
temporal encoding using single networks. Even though available works such as 
C3D+ConvLSTM~\citep{zhu2017multimodal} used two types of networks, the cascaded 
connection makes them dependent on each other during training. How to 
cooperatively train different kinds of networks would be a good research 
direction; for example, using the output of CNN to regularize RNN training 
in parallel.

\textbf{Simultaneous exploitation of spatial-temporal-structural information.} 
A video sequence has three important inherent properties that should be 
considered for motion analysis: spatial information, temporal information and 
structural information. Several previous methods tend to exploit the 
spatio-temporal information for motion recognition, however, structural 
information  contained in the video is rarely  explicitly mined. Concurrent 
mining of these three kinds of information with deep learning would be an 
interesting topic in the future~\citep{jain2016structural}.

\textbf{Fusion of multiple modalities.} While significant progress has been 
achieved by singly using RGB, skeleton or depth modality, effective deep 
networks for fusion of multi-modal data would be a promising direction. For 
example, methods such as SFAM~\citep{Pichaocvpr2017} and 
PRNN~\citep{Shi2017} have pioneered the research in this direction. The work 
SFAM~\citep{Pichaocvpr2017} proposed to extract scene flow for motion 
analysis. The strategy of fusing the RGB and depth modalities at the 
outset allowed the capture of rich 3D motion information. In 
PRNN~\citep{Shi2017} the concept of privileged information (side information) 
was introduced for deep networks training and showed some promise.  
Zolfaghari et al.~\citep{zolfaghari2017chained} proposed the 
integration of different modalities via a Markov chain, which leads
to a sequential refinement of action labels.
So far, most methods considered the three modalities as separate 
channels and fused them at a later or scoring stage using different fusion 
methods without cooperatively exploiting their complementary properties.  
Cooperative training  using different modalities would be a promising research 
area.

\textbf{Large-scale datasets.} With the development of data-hungry deep 
learning approach, there is  demand for large scale RGB-D datasets. Even 
though there are several large datasets, such as NTU RGB+D 
Dataset~\citep{shahroudy2016ntu} and ChaLearn LAP IsoGD 
Dataset~\citep{wanchalearn}, they are focused on specific tasks. Various 
large-scale RGB-D datasets are needed to facilitate research in this field. 
For instance, large-scale fine-grained RGB-D motion recognition datasets and 
large-scale occlusion-based RGB-D motion recognition datasets are urgently 
needed. 

\textbf{Zero/One-shot learning.}  As discussed, it is not always easy to 
collect large scale labeled data. Learning  from a few  examples  remains  a  
key  challenge  in  machine  learning. Despite recent advances in important 
domains such as vision and language, the standard supervised deep learning 
paradigm does not offer a satisfactory solution for learning new concepts 
rapidly from little data. How to adopt deep learning methods for zero/one shot 
RGB-D-based motion recognition would be an interesting research direction. 
Zero/one-shot learning is about being able to recognize gesture/action classes 
that are never seen or only one training sample per class before. This type of 
recognition should carry embedded information universal to all other 
gestures/actions. In the past few years, there are some works on zero/one-shot 
learning. For example, Wan et al.~\citep{pami16Jun} proposed the novel 
spatial-temporal features for one-shot learning gesture recognition and have got 
promising performances on Chalearn Gesture Dataset 
CGD)~\citep{guyon2014chalearn}. For zero-shot learning, Madapana and 
Wachs~\citep{Naveen2017} proposed a new paradigm based on adaptive learning 
which it is possible to determine the amount of transfer learning carried out by 
the algorithm and how much knowledge is acquired for a new gesture observation. 
However, the mentioned works used traditional methods (such as bag of visual 
words model~\citep{jmlr14jun}). 
Mettes et al.~\citep{mettes2017spatial} proposed a spatial-aware 
object embedding for zero-shot action localization and classification. The 
spatial-aware embedding generate action tubes by incorporating word embeddings, 
box locations for actors and objects, as well as their spatial relations. 
However, how to effectively adopt deep learning methods for zero/one 
shot RGB-D based motion recognition would be still an interesting research direction 
especially when using only very few training samples. 

\textbf{Outdoor practical scenarios.} Although lots of RGB-D datasets have been 
collected during the last few years, there is a big gap between the collected 
datasets and wild environment due to constrained environment setting and 
insufficient categories and samples. For example, most available datasets 
do not involve much occlusion cases probably due to the collapse of skeleton 
dataset in case of occlusion. However, in practical scenarios, occlusion is 
inevitable. How to recover or find cues from multi-modal data for such 
recognition tasks would be an interesting research direction. Besides, with the 
development of depth sensors, further distances could be captured, 
and recognition in outdoor practical scenarios will gain the attention of 
researchers.

\textbf{Unsupervised learning/Self-learning.} Collecting labeled datasets are 
time-consuming and costly, hence learning from unsupervised video data is 
required. Mobile robots mounted with RGB-D cameras need to continuously
learn from the environment and without human intervention. How to 
automatically learn from the unlabeled data stream to improve the learning 
capability of deep networks would be a fruitful and useful research 
direction. Generative Adversarial Net (GAN)~\citep{ho2016generative} has 
got much processes recently in image generation task, such as face generation, 
text-to-image task. Besides, it also can be used for recognition task. For 
example, Luan et al.~\citep{tran2017disentangled} proposed a Disentangled 
Representation learning Generative Adversarial Networks (DR-GAN) for 
pose-invariant face recognition. Therefore, we believe the GAN-based techniques 
also can be used for action/gesture recognition, which is a great exciting
direction for research. Carl et al.~\citep{vondrick2016generating} proposed a 
generative adversarial network for video with spatial-temporal convolutional 
architecture that untangles the scene's foreground from backgrounds. This is an 
initial work to capitalize on large amounts of unlabeled video in order to 
learn a model of scene dynamic for both video recognition tasks (e.g. action 
classification) and video generation tasks (e.g. future prediction). Increasing 
research will be reported in the coming years on GAN-based methods for 
video-based recognition.

\textbf{Online motion recognition and prediction.} Online motion recognition and prediction is required in 
practical applications, and arguably this is the final goal of motion 
recognition systems. Differently from segmented recognition, online motion 
recognition requires the analysis of human behavior in a continuous manner, 
and prediction aims to recognize or anticipate actions that would happen. How 
to design effective online recognition and prediction systems with deep 
learning methods has attracted some attention. For example, Vondrick et 
al.~\citep{vondrick2016anticipating} introduced a framework that capitalizes on 
temporal structure in unlabeled video to learn to anticipate human actions and 
objects based on CNN, and it is likely to emerge as an active research area.

\section{Conclusion}\label{conclusion}
This paper presents a comprehensive survey of RGB-D based motion recognition 
using deep learning. We provide a brief overview of existing commonly used 
datasets and pointed at surveys that focused mainly on datasets. The available 
methods are grouped into four categories according to the modality: RGB-based, 
depth-based, skeleton-based and RGB+D-based. 
The three modalities have their own specific features and lead to  
different choices of deep learning methods to take advantages of their 
properties. Spatial, temporal and structural information inherent in a video 
sequence is defined, and from the viewpoint of spatio-temporal-structural 
encoding, we analyse the pros and cons of available methods. Based on the 
insights drawn from the survey, several potential research directions are 
described, indicating the numerous opportunities in this field despite the 
advances achieved to date.

\section*{Acknowledgment}
Jun Wan is partially supported by the National Natural Science 
Foundation of China [61502491]. Sergio Escalera is partially supported by Spanish project [TIN2016-74946-P] (MINECO/FEDER, 
UE) and CERCA Programme / Generalitat de Catalunya.

\bibliographystyle{model2-names}

\end{document}